\pdfoutput=1
\documentclass[11pt]{article}
\usepackage[T2A]{fontenc}
\usepackage[utf8]{inputenc}
\usepackage[russian, english]{babel}

\usepackage[preprint]{acl}

\usepackage{times}
\usepackage[T1]{fontenc}
\usepackage{microtype}
\usepackage{inconsolata}

\usepackage{graphicx}

\usepackage{amsmath}
\usepackage{subcaption}
\usepackage[capitalise]{cleveref}
\usepackage{xcolor}

\newcommand{\Ind}[1]{\mathbf{1}\{#1\}}

\newcommand{\lvar}[2]{\ell^{\mathrm{#1}}_{\mathrm{#2}}}
\newcommand{\linsrc}{\lvar{in}{src}}
\newcommand{\lintgt}{\lvar{in}{tgt}}
\newcommand{\loutsrc}{\lvar{out}{src}}
\newcommand{\louttgt}{\lvar{out}{tgt}}
\usepackage{booktabs}
\usepackage{float}
\usepackage{mdframed}
\graphicspath{{figures_pdf}}

\title{When Meanings Meet: Investigating the Emergence and Quality of \\Shared Concept Spaces during Multilingual Language Model Training}

\author{
 \textbf{Felicia Körner\textsuperscript{1,2}},
 \textbf{Max Müller-Eberstein\textsuperscript{3,4}},
 \textbf{Anna Korhonen\textsuperscript{5}},
 \textbf{Barbara Plank\textsuperscript{1,2}}\\
 \textsuperscript{1}MaiNLP, Center for Information and Language Processing, LMU Munich, Germany\\
 \textsuperscript{2}Munich Center for Machine Learning (MCML), Munich, Germany\\
 \textsuperscript{3}University of Tokyo, Japan\\
 \textsuperscript{4}IT University of Copenhagen, Denmark\\
  \textsuperscript{5}Language Technology Lab, University of Cambridge, United Kingdom\\
 \small{
   \textbf{Correspondence:} \href{mailto:f.koerner@lmu.de}{f.koerner@lmu.de}
 }
}

\begin{document}
\maketitle
\begin{abstract}
Training Large Language Models (LLMs) with high multilingual coverage is becoming increasingly important\textemdash especially when monolingual resources are scarce. Recent studies have found that LLMs process multilingual inputs in shared concept spaces, thought to support generalization and cross-lingual transfer. 
However, these prior studies often do not use causal methods, lack deeper error analysis or focus on the final model only, leaving open how these spaces emerge \emph{during training}.
We investigate the development of language-agnostic concept spaces during pretraining of EuroLLM through the causal interpretability method of activation patching. We isolate cross-lingual concept representations, then inject them into a translation prompt to investigate how consistently translations can be altered, independently of the language.
We find that \emph{shared concept spaces emerge early} and continue to refine, but that \emph{alignment with them is language-dependent}.
Furthermore, in contrast to prior work, our fine-grained manual analysis reveals that some apparent gains in translation quality reflect shifts in behavior\textemdash like selecting senses for polysemous words or translating instead of copying cross-lingual homographs\textemdash rather than improved translation ability. Our findings offer new insight into the training dynamics of cross-lingual alignment and the conditions under which causal interpretability methods offer meaningful insights in multilingual contexts.
\end{abstract}
\begin{figure}[t!]
  \centering
  \includegraphics[width=0.90\linewidth]{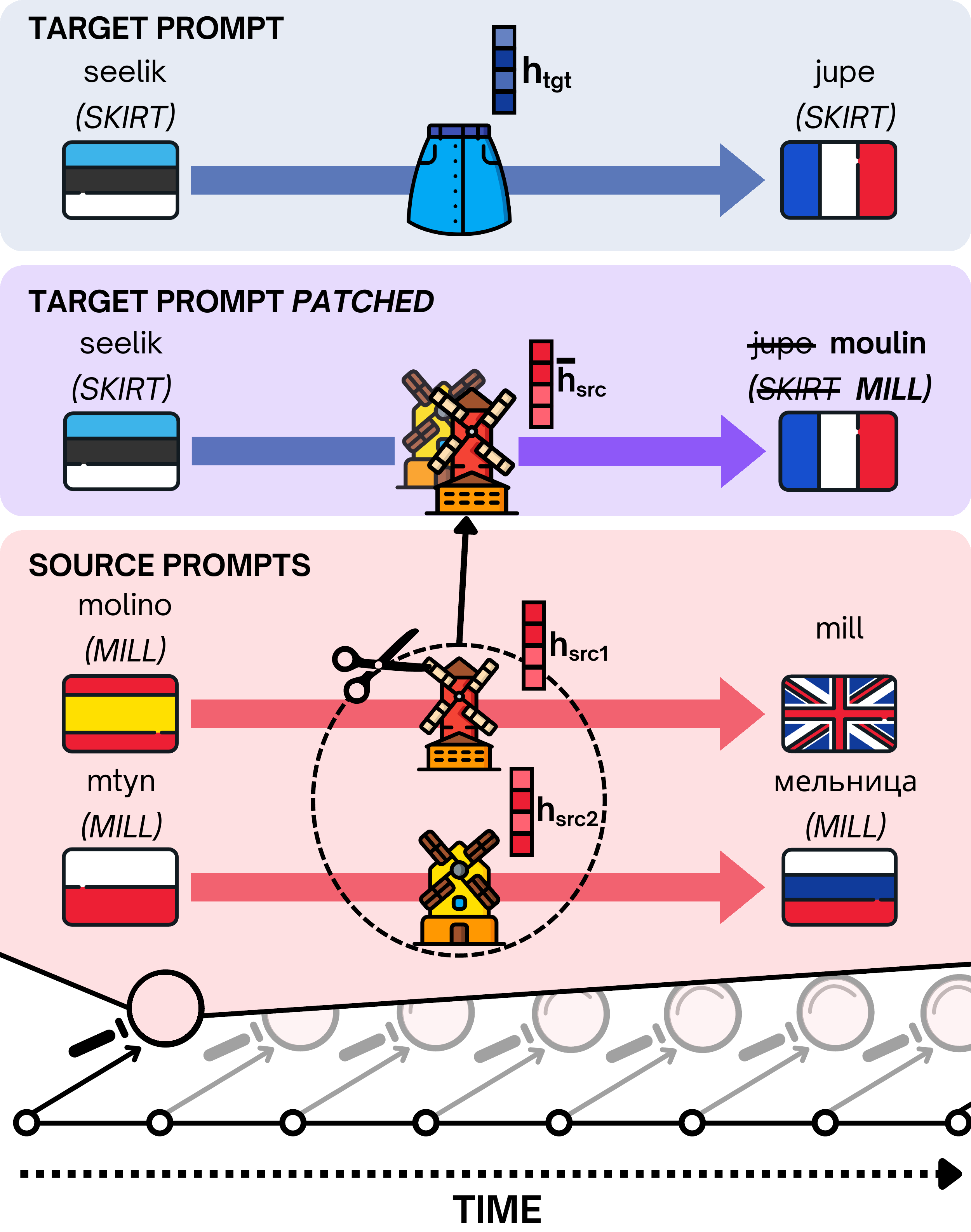}
  \caption{We extend cross-lingual concept patching  \cite{dumas-etal-2025-separating} through a systematic application over pretraining. The first row shows the vanilla translation of the concept $\textsc{skirt}$ from Estonian $\rightarrow$ French. In the second, we replace the activation with averaged activations from Spanish $\rightarrow$ English and Polish $\rightarrow$ Russian for $\textsc{mill}$. Though the representation comes from a \emph{distinct} set of languages, it can induce $\textsc{mill}$ in French.}
  \label{fig:patching}
\end{figure}
\section{Introduction}
Most Large Language Models (LLMs) are trained primarily on English, and even targeted multilingual training is typically imbalanced across languages \cite{NEURIPS2024_1bd359b3, Liu2024ResponsibleMLA}. Despite this, LLMs exhibit emergent cross-lingual alignment, enabling 
transfer of capabilities to other languages \cite{chirkova-nikoulina-2024-zero-shot}.

Since data is scarce for many languages \cite{joshi-etal-2020-state, ranathunga-de-silva-2022-languages}, this behavior is critical for narrowing the performance gap between English and other languages and to develop more inclusive language technologies \cite{Peppin2025TheMD, li-etal-2024-prealign}.
Understanding how and \emph{when} cross-lingual alignment arises is crucial for determining the necessary factors for training multilingually performant LMs.

Recent interpretability work has found that models rely on shared concept\footnote{We use the term ``concept'' to describe a mental representation expressed by different words across languages.} spaces to process multilingual input \cite{wendler-etal-2024-llamas}. To causally demonstrate the existence of such spaces \citet{dumas-etal-2025-separating} propose \emph{cross-lingual concept patching} (see \cref{fig:patching} for an illustration). In cross-lingual concept patching, latent vectors are extracted from a prompt translating one concept (e.g.\ $\textsc{mill}$ in \cref{fig:patching}) across one set of language pairs (e.g.\ Spanish $\rightarrow$ English and Polish $\rightarrow$ Russian) and artificially applied during the translation of a \emph{different} concept (e.g.\ $\textsc{skirt}$) across a \emph{different} language pair (e.g.\ Estonian $\rightarrow$ French). If this patched inference produces the first concept ($\textsc{mill}$) in the new language (French), this provides causal evidence for shared concept representations, as language-specific ones would not be able to change the concept while keeping the output language fixed.

In this work, we re-frame cross-lingual concept patching as a causal analysis of cross-lingual alignment. The key intuition is that \emph{two factors} are involved in successfully inducing a concept change from concept representations averaged across languages: First, the languages the representations are drawn from must be aligned with the shared space. Second, the output language must be aligned with these representations, even when they stem from other languages.

However, it is not obvious how shared spaces develop, and how models learn to map them to a particular output language\textemdash a gap we aim to address in this study. One hypothesis is that models initially rely on language-specific representations and only gradually abstract these into a shared space over time. To test this, we apply cross-lingual concept patching across intermediate checkpoints during pretraining. Specifically, our contributions are:
\begin{itemize}
\item We introduce a systematic framework for cross-lingual concept patching: we carefully curate compatible source and target concept pairs (\cref{sec:related-work} and \cref{sec:data}) and devise distinct language settings, including a control task (\cref{sec:lang-settings}).
We find that \textbf{alignment with shared spaces depends on training data proportion}.
\item We conduct the first fine-grained investigation into the emergence of shared concept spaces during multilingual pretraining, studying intermediate checkpoints of EuroLLM, which provides a higher granularity of checkpoints than previously-studied multilingual LLMs (\cref{sec:results}). We find that \textbf{shared concept spaces arise early in pretraining}.
\item We apply cross-lingual concept patching to the largest and most diverse set of concepts to date (\cref{sec:data}).
\item We provide the first manual analysis of errors under cross-lingual concept patching (\cref{sec:manual-eval}), uncovering nuance in its failure modes, and interpretation of its effect.
\end{itemize}
We share our implementation to facilitate further research.\footnote{ \url{https://github.com/mainlp/shared-concept-spaces}}

\section{Related Work}\label{sec:related-work}
Recent mechanistic interpretability work has proposed that multilingual LLMs process input in three stages: i) mapping from the input language to a shared concept space, ii) processing conceptual information, and iii) mapping the result to the output language \cite{NEURIPS2024_1bd359b3, schut2025do, wendler-etal-2024-llamas, tezuka-inoue-2025-transfer, zhong-etal-2025-language}.

In a similar vein, several studies have identified shared components of multilingual LLMs, largely through neuron analysis \cite{stanczak-etal-2022-neurons, 10.1609/aaai.v38i16.29735, DBLP:journals/corr/abs-2411-17401, tang-etal-2024-language, Zhang2025HowDA, DBLP:journals/corr/abs-2406-09265, kojima-etal-2024-multilingual}, though often not causally.
Recent work highlights the importance of causal interventions for robust mechanistic claims \cite{DBLP:journals/corr/abs-2408-01416}, a perspective we adopt here. Methodologically, \citet{dumas-etal-2025-separating} is closest to our work; they introduce cross-lingual concept patching, causally demonstrating the existence of language-agnostic concept spaces. \citet{feucht2025the} use the same dataset to show that concept induction heads, used for copying concepts mono- and cross-lingually, are language-agnostic. Both works use the same method to demonstrate shared components at different granularity, but rely on a dataset of constructed concept pairs via LLM translation. 
In contrast, we re-frame the method to \emph{analyze} how shared spaces develop. To do so, we introduce a systematic framework to compare results across different language settings and pretraining checkpoints, and propose a novel, carefully curated concept dataset extracted from human translations.

Despite growing evidence of shared spaces in multilingual models, little is known about how such spaces emerge throughout pretraining. Progress is hampered by the limited availability of pretraining checkpoints for state-of-the-art multilingual LLMs. Until the recent release of Apertus \cite{DBLP:journals/corr/abs-2509-14233}, BLOOM \cite{workshop2023bloom176bparameteropenaccessmultilingual} was the only model with publicly available checkpoints, yet provides only 4–8,\footnote{Depending on model size.} potentially obscuring finer-grained shifts. Studies of BLOOM use neuron analysis, probing, and, in some cases, concept-level analysis \cite{zeng-etal-2025-converging, wang-etal-2024-probing-emergence, riemenschneider-frank-2025-cross}.
In contrast, we study 26 pretraining checkpoints of EuroLLM, and provide additional results for Apertus 8B and OLMo-2 7B in \cref{sec:other-models}.

\section{Methodology}\label{sec:method}
We study model outputs under cross-lingual concept patching \cite{dumas-etal-2025-separating}, an activation patching \cite{10.5555/3495724.3496763} approach. Activation patching is a mechanistic interpretability method, whereby activations of a forward pass of a ``target'' prompt are overwritten, or ``patched'', with activations from a previous run of a different, ``source'', prompt. The effect of this intervention provides insight about the role of the patched component and the patch itself. We describe the variant of cross-lingual concept patching in the following subsections, and give more technical details in \cref{sec:tech}. 

\subsection{Few-shot Concept Translation}
As a testbed for cross-lingual concept patching, we use the task of few-shot, word-level translation.
The task considers the translation of a concept $C$
from an input language, $\ell^{\mathrm{in}}$, to an output language, $\ell^{\mathrm{out}}$.
We denote a language-agnostic concept with capital letters, for example $C=\textsc{mill}$. $C^{\ell^{\mathrm{out}}}$ is the concept expressed as a word in the output language, for example $C^{FR}=$``moulin''.

Each translation is formulated as a prompt in the format, ``$\ell^{\mathrm{in}}$: $C^{\ell^{\mathrm{in}}}$ - $\ell^{\mathrm{out}}$: $C^{\ell^{\mathrm{out}}}$'', where the output concept is omitted during inference. The prompt is further prepended with five few-shot examples. For example, a translation prompt from Spanish to English for $C=\textsc{mill}$:
\begin{mdframed}[backgroundcolor=red!5, linecolor=red, linewidth=0.5pt]
Español: ``cultura" - English: ``culture''\\
Español: ``sentido" - English: ``sense''\\
...\\
Español: ``molino" - English: ``
\end{mdframed}
\subsection{Cross-lingual Concept Patching}\label{sec:patch}
To induce a change in translation, we apply a concept patching intervention during inference (see \cref{fig:patching} for an illustration). First, we compute a ``patch'', an activation for $C_{\mathrm{src}}$, the source concept that should be generated.
This activation is extracted from a few-shot translation prompt in which $C_{\mathrm{src}}$ is the intended translation at the position of the last token of the word to be translated for some layer $j$ and all subsequent ones. Extending our previous example, $C_{\mathrm{src}}=\textsc{mill}$, $\linsrc=$ Spanish, and $\loutsrc=$ English, and we extract the activation at the last token of \textcolor{red}{``molino''}. We repeat this for several source language pairs, and take the mean of the activations.

Next, we prompt the model to translate a semantically distinct target concept $C_{\mathrm{tgt}}$ across an entirely different target language pair. E.g., for $C_{\mathrm{tgt}}=\textsc{skirt}$, $\lintgt$= Estonian, and $\louttgt=$ French:

\begin{mdframed}[backgroundcolor=blue!5, linecolor=blue, linewidth=0.5pt]
Eesti: ``korporatsioon" - Français: ``fraternité''\\
Eesti: ``lambatall" - Français: ``agneau''\\
...\\
Eesti: ``seelik" - Français: ``
\end{mdframed}

However, during inference of this target prompt, we artificially apply the averaged concept activations from the source prompts, at the equivalent position, and all subsequent layers. The goal of the intervention is to induce the source concept ($\textsc{mill}$) in the target output language, French, even if the target prompt aims to translate the target concept ($\textsc{skirt}$). I.e., in our example:
\textcolor{blue}{``Eesti: ``seelik" - Français: ``}\textcolor{red}{moulin''}.

Crucially, since the source languages are distinct from the target languages, and the source concept is distinct from the target concept, successfully inducing the source concept in the target language hinges on two representational criteria: First, concept representations must be \textit{language-agnostic} in order for the mean over representations across languages to be meaningful. Second, the target output language must also be \textit{aligned} with this language-agnostic space in order for the model to generate the source concept in the target output language.

\section{Experimental Setup}
\subsection{Model}\label{sec:ellm}
To investigate the emergence of shared concept spaces throughout training, and to link them to concrete training strategies, we analyze 26 pretraining checkpoints of the open-weight EuroLLM-1.7B (\citealp{martins2024eurollmmultilinguallanguagemodels}, henceforth EuroLLM). EuroLLM is trained primarily on European languages; we categorize our languages of focus based on their proportion in EuroLLM's training data in \cref{tab:lang-data}.\footnote{Cantonese is given a special category, though it is not included in EuroLLM's training data, the model translates it fairly well (\cref{sec:results-word-level}). We attribute this to Cantonese's similarity to Mandarin.} The model is of particular interest to our study, as it offers a high granularity of intermediate checkpoints, and is trained in \emph{two phases} with different multilingual training data compositions.

Specifically, its training data is composed of web data, code/math data, high-quality data (Wikipedia, arXiv, books, medical texts), and parallel data in two different configurations. In phase one (0--90\% of training; 3.6T tokens), 77\% of the data is \emph{web}, and \emph{parallel data} is primarily aligned with respect to English. In phase two (90\%--100\% of training; 0.4T tokens), \emph{web} is reduced to 46.6\%, while the \emph{high-quality data} ratio is increased and upsampled from 9\% to 34.4\%. The amounts of \emph{code/math} and \emph{parallel data} remain similar, however, the parallel data is drawn from multi-aligned sources beyond English. As such, it is of particular interest to investigate how each training phase manifests in terms of cross-lingual concept alignment.

\begin{table}[t]
\centering
\resizebox{\columnwidth}{!}{%
\begin{tabular}{@{}lll@{}}
\toprule
\textbf{Category} & \textbf{Languages} & \textbf{Training Data Proportion} \\
\midrule
very-high         &  \texttt{en}            & 50\% phase one, 32.5\% phase two       \\
high               & \texttt{es}, \texttt{fr}     & around 6\%                         \\
med-high           & \texttt{zh}            & between 3--4\%                     \\
med                & \texttt{ru}, \texttt{pl}      & between 2--3\%                     \\
low                & \texttt{et}, \texttt{fi}      & around 1\%                         \\
unseen but similar & \texttt{yue}          & --               \\
unseen             & \texttt{sw}, \texttt{cy}     & --              \\
\bottomrule
\end{tabular}%
}
\caption{Our languages of focus, selected from Multi-SimLex (\cref{sec:data}) and categorized based on EuroLLM's training data proportions (\cref{sec:ellm}).}
\label{tab:lang-data}
\end{table}
\subsection{Data}\label{sec:data}
To curate our dataset, we leverage Multi-SimLex \cite{vulic-etal-2020-multi}
as it provides a large, linguistically informed and diverse concept set in contrast to prior work, which focused on synthesized ``picturable'' concepts only \cite{feucht2025the, dumas-etal-2025-separating}. Multi-SimLex consists of 1,888 word pairs rated for lexical similarity, with human translations in 13 languages. The languages are typologically diverse, spanning both low- and high-resourced languages.
We focus on the following 11 languages, covering eight languages included and three excluded from EuroLLM’s training data (\cref{sec:ellm}): English (\texttt{en}), Mandarin (\texttt{zh}), Welsh (\texttt{cy}), Estonian (\texttt{et}), Finnish (\texttt{fi}), French (\texttt{fr}), Polish (\texttt{pl}), Russian (\texttt{ru}), Spanish (\texttt{es}), Swahili (\texttt{sw}), Cantonese (\texttt{yue}). Multi-SimLex was originally created for word-pair similarity judgments, however, we do not retain the original word pairings. Instead, we treat the dataset as a multiway parallel lexicon, extracting a vocabulary of 2,147 concepts and their translations. Prior work has shown that word classes may behave differently\textemdash for example, function words are often language-specific \cite{schut2025do}. To avoid confounding effects of word class, and due to the extensive scope of our experiments across languages and checkpoints, we focus on a single class and follow prior work \cite{dumas-etal-2025-separating, feucht2025the} in restricting analysis to nouns.

We greedily select 256 \emph{compatible} source and target concept pairs from this pool, where we deem a pair compatible if there is no word overlap across all translations for the two concepts, for a total of 398 distinct concepts. For each source concept, we construct source prompts across all source language pairs, and for each target concept we construct target prompts across all target language pairs. So far, cross-lingual concept patching has been applied to 200 pairs constructed from a pool of about 100 concepts \cite{feucht2025the,dumas-etal-2025-separating}.
\begin{figure*}
\centering
\includegraphics[width=\linewidth, height=0.45\textheight, keepaspectratio]{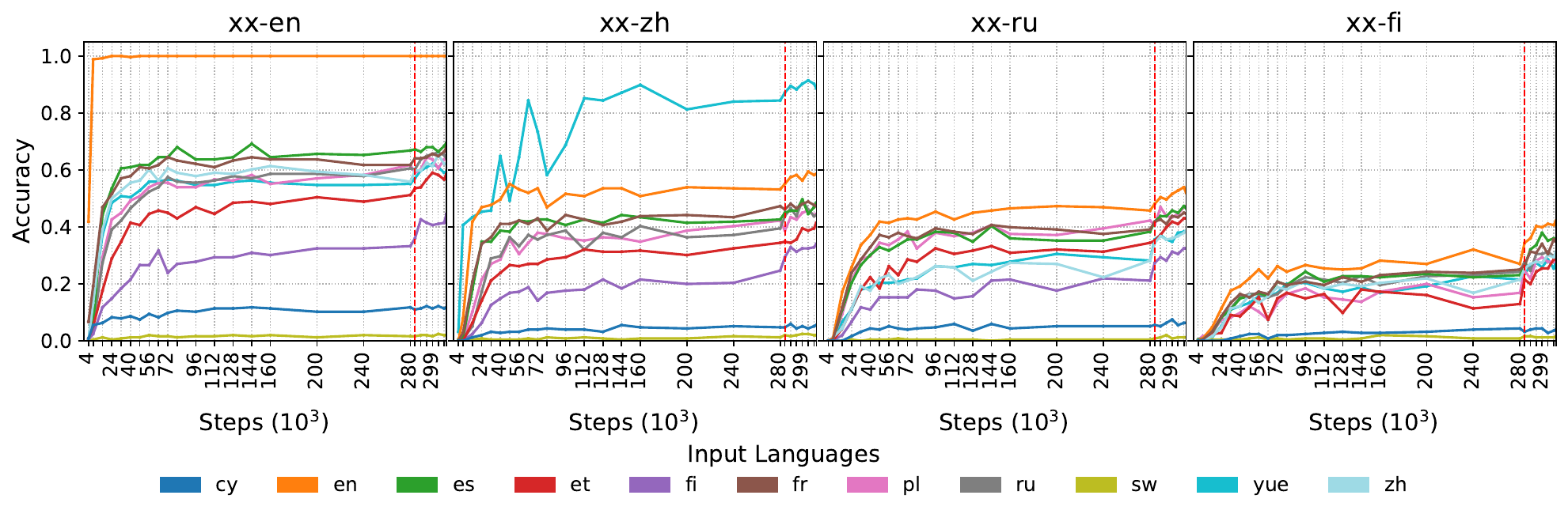}
\caption{Mean word-level translation accuracy over checkpoints for source prompts used for patching, grouped by a selection of output languages. The red dotted line indicates the start of phase two of EuroLLM's training.}
\label{fig:translation}
\end{figure*}

 \subsection{Evaluation}\label{sec:eval}
Activation patching evaluations typically measure the increase in next-token probability for the expected first token \cite{DBLP:journals/corr/abs-2404-15255}. However, as our results in the word-level translation experiments in \cref{sec:results-word-level} reveal, this metric is poorly suited for our task, where there are often multiple valid next-token sequences for a given concept. Unlike prior work, we do not expand the dataset with synonyms for evaluation, as we favor a precise fine-grained analysis, while covering a more diverse set of concepts and study of lower-resourced languages (see \cref{sec:babelnet}). Finally, relying on the proxy of first-token probability may mask errors where the first-token appears correct, but the model outputs the wrong prediction. For example, consider ``organ'' and ``organizer'', which may share a first token, but are semantically distinct. To address these limitations, we evaluate the model on full token sequences and measure word-level translation accuracy.

In particular, given $N$ test samples, $y_i$ the \emph{source} concept expressed in the \emph{target} output language, and $\hat y_i$ the model prediction, we define mean word-level translation accuracy as:
\[
\mathrm{Acc}=\frac{1}{N}\sum_{i=1}^{N}\Ind{\hat y_i = y_i}.
\]
This approach directly assesses whether patching meaningfully alters the translation, and furthermore, avoids cases where the first token appears correct but the final prediction is wrong. Finally, we conduct a multilingual manual evaluation (\cref{sec:manual-eval}) to gain a deeper understanding of concept translation performance.

\subsection{Language Settings}\label{sec:lang-settings}
We use all languages in \cref{tab:lang-data} except \texttt{cy} and \texttt{yue} as target input languages, treating \texttt{sw} as representative for unseen languages. We fix the target output languages to \texttt{en}, \texttt{ru}, and \texttt{zh}, covering three typologically distinct languages with diverse scripts.

For each target language pair ($\lintgt, \louttgt$), we define three sets of source language pairs used for patching: \texttt{seen}, \texttt{en\_en} and \texttt{tgt}. \texttt{seen} is made up of languages in EuroLLM's training data, excluding \texttt{en} and the languages in the target pair. For example, for target pair \texttt{fr--en}, \texttt{seen} includes all ordered pairs of \texttt{es}, \texttt{zh}, \texttt{pl}, \texttt{ru}, \texttt{et}, \texttt{fi}, \texttt{yue}, for 42 total source language pairs.

\texttt{en\_en} is a copying task, formulated in the same way as the translation task, except that both the input and output language are \texttt{en}. We include this as a control task and a strong baseline, reasoning that the \texttt{en\_en} representations will be well-defined as English dominates EuroLLM's training data. \texttt{tgt} is an ablation, where only the source and target concepts differ, and the source language pair matches the target language pair. Both \texttt{en\_en} and \texttt{tgt} consist of only a single language pair, hence, there are no concept-aligned representations to average over. 

Given \texttt{msimlex} as the set of selected languages (\cref{sec:data}), \texttt{seen}, \texttt{en\_en} and \texttt{tgt} are defined:
{
\begin{align*}
\texttt{ellm} &= \texttt{msimlex} \setminus \{\texttt{sw},\ \texttt{cy}\} \\
\texttt{seen} &= \{(\ell_1,\ell_2) \mid \ell_1 \neq \ell_2,\\
&\quad\quad \ell_1,\ell_2 \in \texttt{ellm} \setminus \{\lintgt,\ \louttgt,\ \texttt{en}\}\}\\
\texttt{en\_en} &= \{ (\texttt{en},\ \texttt{en}) \}\\
\texttt{tgt} &= \{ (\lintgt,\ \louttgt) \}
\end{align*}
}

As a baseline, we evaluate the unpatched translation of the source concept for the target language pair. I.e., if we aim to induce the source concept $\textsc{mill}$ for the target language pair \texttt{et--fr} through cross-lingual patching, we compare this to the vanilla translation of $\textsc{mill}$ for \texttt{et--fr}. We denote this as \texttt{src\_unpatched}. Please see \cref{sec:example-prompts} for example prompts to illustrate these settings.

\section{Results}\label{sec:results}
\begin{figure*}
\centering
\includegraphics[width=\linewidth, height=0.4\textheight, keepaspectratio]{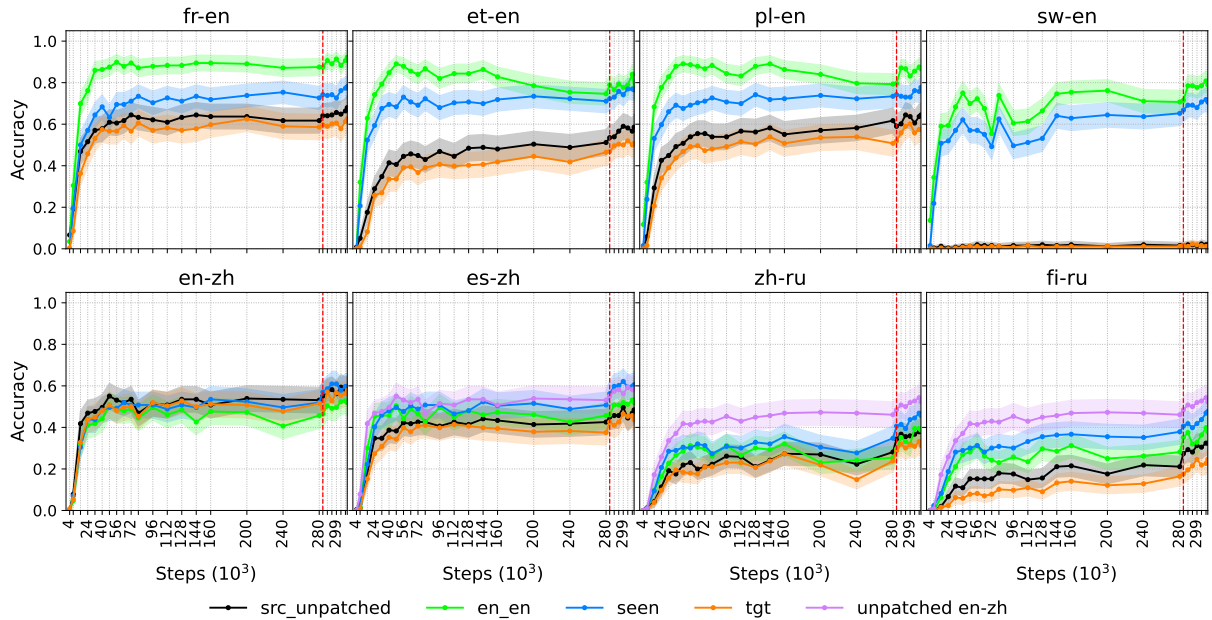}
\caption{Mean word-level translation accuracy over checkpoints under different patching settings for a selection of target language pairs. We overlay \texttt{en--xx}, where \texttt{xx} is \texttt{zh} or \texttt{ru} on \texttt{es--zh}, and \texttt{fi--ru}, \texttt{zh--ru}, respectively. The red dotted line indicates the start of phase two of EuroLLM's training. We show 95\% CI over 256 samples.}
\label{fig:obj-patch-merged}
\end{figure*}
\subsection{Word-Level Translation}\label{sec:results-word-level}
We first evaluate EuroLLM's checkpoints on the unpatched word-level translation task to confirm our assumptions about the relationship between training data proportion (\cref{tab:lang-data}) and translation performance. Unsurprisingly, we find that translation accuracy (see \cref{fig:translation} for a selection of languages, \cref{fig:full-translation} in the App.\ for all) correlates with training data proportion. The model largely fails to translate words where the input or the output language is unseen, namely, \texttt{cy} and \texttt{sw}. The exception is \texttt{yue}, which is well-translated, in particular to or from \texttt{zh}, likely due to its similarity to \texttt{zh}.

We interestingly observe that translation accuracy improves in the second phase of training, in particular for lower-resourced languages, such as \texttt{pl}, \texttt{ru}, \texttt{fi}, and \texttt{et}. We hypothesize that the multiway parallel data introduced in this phase helps align these languages, improving translation quality \cite{shen-etal-2025-unaligned, lin-etal-2025-recipe}.

The \texttt{en\_en} curve shows that the model quickly becomes proficient at copying input words\textemdash by the second checkpoint words are consistently copied. In fact, the model appears to prefer copying over translating in the early stages, even for the translation task (i.e., when processing the prompts for language pairs other than \texttt{en\_en}). This behavior is in line with recent work suggesting that copying is learned before other tasks \cite{feucht2025the}.

\paragraph{Manual Analysis of Unpatched \texttt{fr--en}}\label{sec:en-fr}
Using \texttt{fr--en} as a case study, we inspect the translations and find that many apparent errors can be attributed to dataset artifacts or model behavior. In particular, our analysis sheds light on the model's copying behavior. In some cases, these are loanwords, hence, the translation could be considered correct. E.g., when translating ``balustrade'', where the expected translation is ``rail'', the model consistently outputs ``balustrade''. However, cross-lingual homographs are also copied, resulting in an incorrect translation. For example, the model copies both ``course'' (expected: ``racing''), and ``coin'' (expected: ``corner'').

Similarly, the results are confounded by polysemous words, which are also not captured by the translation dataset. For example, the model translates ``femme'' (expected: ``wife'') to ``woman'', or ``pécheur'' (expected: ``sinner'') to ``fisherman''. These factors motivate the manual analysis of errors under the \texttt{seen} patching setting (\cref{sec:manual-eval}), to distinguish actual errors from reasonable outputs which are not captured by the automatic evaluation.

\subsection{Cross-lingual Concept Patching}\label{sec:results-patching}
\begin{figure}
\centering
\includegraphics[width=\linewidth]{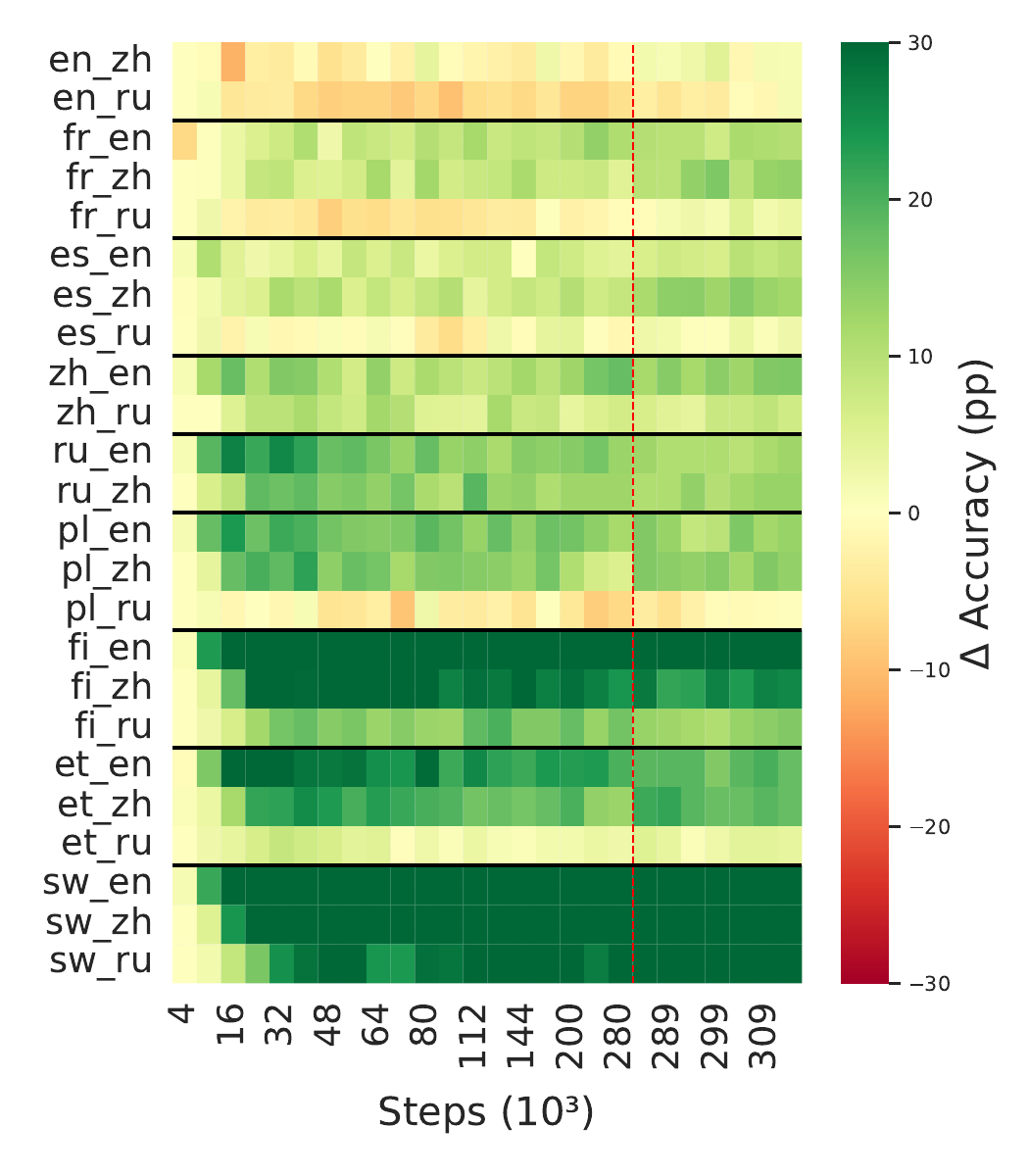}
\caption{Net improvement (in percentage points, capped at $\pm35$pp) of \texttt{seen} over unpatched translation accuracy. Each row shows a target language pair, grouped by input and ordered by output language; each column one checkpoint. The red dotted line indicates the start of phase two of EuroLLM's training.}
\label{fig:net}
\end{figure}

Patching from \texttt{seen} results in comparable or better accuracy than the unpatched translation across checkpoints for most target language pairs (\cref{fig:net}), providing strong evidence that \textbf{language-agnostic spaces arise early in pretraining}.

\paragraph{Effect of Source Language Pair Groups}
\texttt{en\_en} is initially the strongest patching setting across all target language pairs with \texttt{en} as the output language (\cref{fig:obj-patch-merged}, see App.\ \cref{fig:obj-patch-full-en}, \cref{fig:obj-patch-full-zh}, \cref{fig:obj-patch-full-ru} for full results for \texttt{en}, \texttt{zh}, \texttt{ru}, respectively). This aligns with our expectation that \texttt{en\_en} provides well-aligned concept representations and the model is proficient at mapping these to \texttt{en} output. However, \texttt{seen} does not lag far behind. For target language pairs with \texttt{zh} or \texttt{ru} as the target output language, \texttt{en\_en} is consistently on par with or \emph{worse} than \texttt{seen}. This is surprising, as we expect both \texttt{zh} and \texttt{ru} to be well-aligned with \texttt{en}. In fact, unpatched \texttt{en--ru} and \texttt{en--zh} are an upper bound for the patched translation accuracy for target output \texttt{ru} and \texttt{zh} respectively. Overall, the similar behavior under the \texttt{en\_en} and \texttt{seen} settings across checkpoints suggests that language-specific concept spaces do not strongly precede the emergence of shared spaces.

As an additional observation, accuracy under our ablation setting, \texttt{tgt}, is comparable to or lower than the unpatched translation, particularly for lower-resourced languages. Thus, when source and target language pairs are identical, patching results in a lower accuracy for the induced source concept, suggesting that information from the forward pass of the target prompt is not entirely overwritten.

\paragraph{Effect of Target Language Pair}
Under patching, the input language has much less of an effect on performance than in the unpatched setting. This aligns with recent work on multilingual processing, which suggests that models map from the input language to a shared space before mapping back to the output language \cite{wendler-etal-2024-llamas,NEURIPS2024_1bd359b3}. Since we patch at intermediate layers, the model's ability to map input from each language to the shared concept space becomes less important. Nevertheless, the input language does play a role. This is most visible for the target pair \texttt{sw--en}, where accuracy fluctuates significantly more than for other language pairs. \cref{fig:net} shows the impact of the target output language; the lower-resourced output language \texttt{ru} sees less gains from patching than \texttt{zh}, which, in turn, sees less gains than \texttt{en}. We hypothesize this reflects how well each output language is aligned with shared spaces, in accordance with theory on multilingual processing.
\begin{figure*}
\centering
\includegraphics[width=\linewidth, height=\textheight, keepaspectratio]{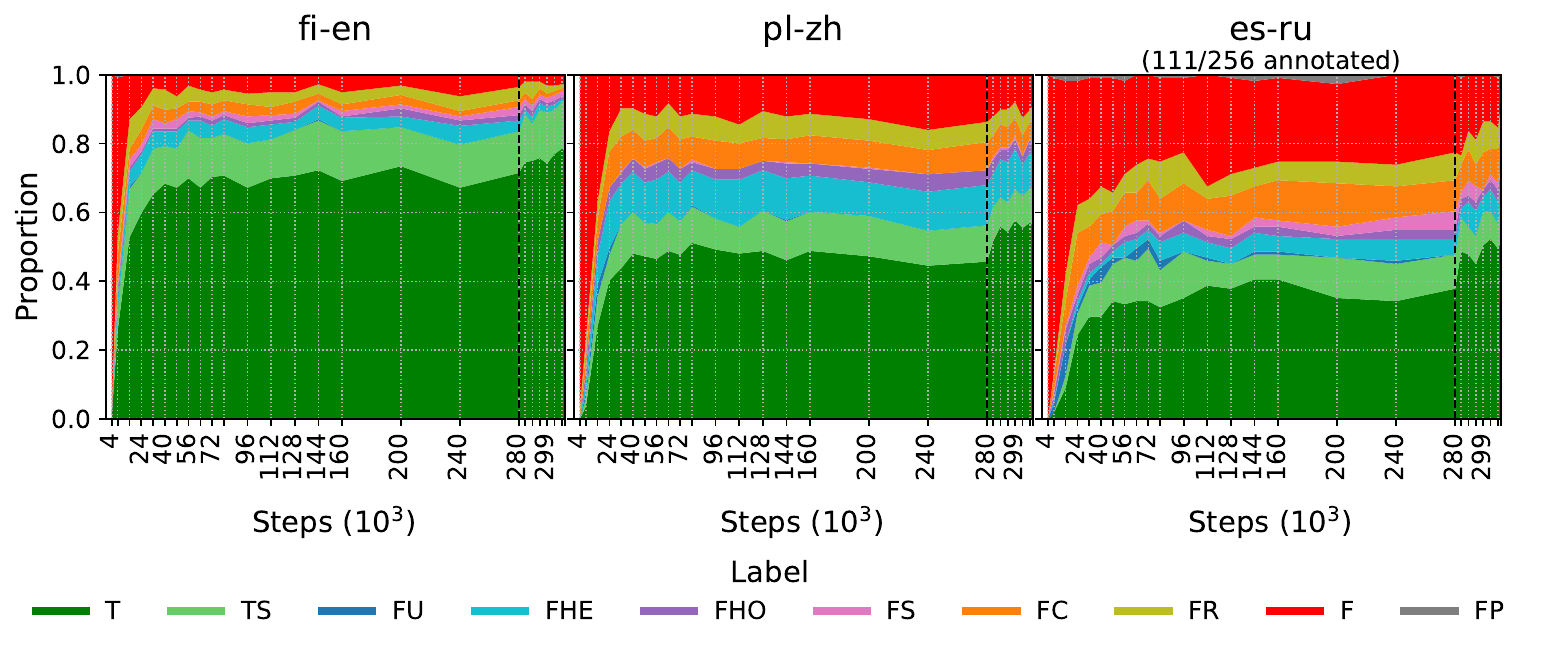}
\caption{Area grid of label distribution for outputs under the \texttt{seen} patching setting for a selection of languages. For \texttt{ru} we show only the 111 concepts for which outputs that were labeled; \texttt{en} and \texttt{zh} were labeled in full. The black dotted line indicates the start of phase two of EuroLLM's training.}
\label{fig:labels}
\end{figure*}
\paragraph{Concept-Specific Effects}
Since more frequent words have been found to be better aligned across languages \cite{peng-sogaard-2024-concept}, we analyze the effect of frequency on translation behavior under patching. We focus on the target output language \texttt{en}; assuming that frequency estimates derived from English fastText embeddings \cite{grave-etal-2018-learning}\footnote{cc.en.300.bin} provide a better approximation of EuroLLM’s token frequency distribution than analogous estimates for \texttt{ru} or \texttt{zh}, which are less represented in the pretraining corpus. 

Under the \texttt{seen} setting, the model outputs the target word more often than a synonym for more frequent target words. This likely reflects the relative frequency of the synonym to the target word, and appears to hold for the unpatched setting, though less clearly. The most frequent words are mostly translated correctly early on, and only occasionally regress. For less frequent targets, the patched translation is more often incorrect, with some almost never translated correctly. The relationship between frequency and translation quality is far less pronounced for the unpatched case. These observations hold across target language pairs with output language of \texttt{en} (see \cref{fig:seen-unpatched-freq} in the App.\ for categorical maps for \texttt{es--en}, \texttt{fr--en}, \texttt{zh--en}, \texttt{ru--en}).

Notably, for \texttt{fr} and \texttt{es} patching yields more incorrect translations for the least frequent concepts compared to unpatched. We attribute this to stronger unpatched translation for higher‑resourced languages and poorly-aligned representations for rare concepts, making patching less effective.
\subsection{Manual Error Analysis of \texttt{seen} Setting}\label{sec:manual-eval}
\begin{table}[ht]
\centering
\resizebox{\columnwidth}{!}{%
\begin{tabular}{lll}
\toprule
\textbf{Label} & \textbf{Description} & \textbf{Example}\\
\midrule
\texttt{T}    & exact match (automatic) & poverty $\rightarrow$ poverty \\
\midrule
\texttt{TS}   & synonym &  outlander $\rightarrow$ foreigner \\
\midrule
\midrule
\texttt{FS}   & differ in one aspect of grammar & outlander $\rightarrow$ foreigners \\
&  to a T or TS & actor $\rightarrow$ actress \\
\midrule
\texttt{FHE}  & hypernym relative to target & liquor $\rightarrow$ drink \\
\midrule
\texttt{FHO}  & hyponym relative to target & insect $\rightarrow$ honeybee \\
\midrule
\texttt{FR}   & co-hyponym to target & skirt $\rightarrow$ dress \\
\midrule
\texttt{FU}   & untranslated & culture $\rightarrow$ cultura \\
\midrule
\texttt{FC}   & conceptually similar to target & archer $\rightarrow$ arrow \\
\midrule
\texttt{F}    & wrong & aunt $\rightarrow$ sheep\\
\bottomrule
\end{tabular}
}
\caption{Label definitions for manual error analysis. Labels are mutually exclusive, and ordered by priority, i.e.\ \texttt{FS} overrides \texttt{FC}; \texttt{F} only applies if no other labels do (see \ref{fig:decision-tree} in the App.\ ) \texttt{T} and \texttt{TS} are considered correct.}
\label{tab:categories}
\end{table}

To better understand the quality and evolution of the shared concept representations, and how the model fails to map them to the expected output, we conduct a novel error analysis. Specifically, outputs under \texttt{seen} are annotated by native speakers using the labels we propose in \cref{tab:categories}. Due to resource constraints, we annotate a subset of outputs for \texttt{ru}, corresponding to 111 concepts (see \cref{sec:annotation} for more information on annotation). Results are provided in \cref{fig:labels}.

We consider the categories \texttt{F*} (excluding \texttt{F}) ``partially'' correct. For all language settings, outputs are more often partially correct than fully wrong, and the proportion of \texttt{F} stays mostly stable, or decreases. This is evidence that the averaged concept representations already encode conceptual and syntactic information prior to inducing the desired concept, and that they are refined throughout training.

We do not observe language-specific patterns on the target input side. However, the distribution of categories is dependent on the output language, e.g.\ \texttt{FS} makes up a larger proportion of the errors for \texttt{ru}, which likely reflects its complex morphology.

In general, \texttt{FHO} is less common than \texttt{FHE}, aligning with our intuition that poorly-aligned concept spaces are more likely to be mapped to general terms. We also note that \texttt{FR} is a relatively large class, which may reflect the language model training objective, i.e., the model learns that a word plays a particular syntactic role.

Interestingly, for all target language pairs, in phase two the proportion of \texttt{T} increases more than \texttt{TS} (which remains stable),
suggesting increased cross-lingual alignment through parallel data, resulting in more specific representations that induce the target word rather than a synonym.

Linking back to errors seen in \cref{sec:en-fr}, we find that patching shifts sense selection for polysemous words. For example, for \texttt{fr--en} `avocat' may be translated to `attorney` (expected) or `avocado'. Under \texttt{seen} patching, the model output is shifted from `avocado' to `attorney'. For \texttt{es}-\texttt{en}, `mañana' is translated to `morning' in the unpatched case and `tomorrow' under patching. Since we average over representations for many language pairs, polysemy for particular pairs is lost, leading to a selection of the dominant shared sense. Similarly, patching discourages the model from copying cross-lingual homographs. For example, for \texttt{fr--en}, the model copies `tour' in the unpatched case, whereas under patching, the model outputs the expected `tower'.
\section{Conclusion}
We present the first fine-grained investigation into the emergence of shared concept spaces during multilingual pretraining, showing that shared concept spaces \emph{emerge early and remain relatively stable throughout training}. Our novel dataset and error analysis reveals that intermediate concept representations encode meaningful semantic information even before they can be mapped to specific concepts, suggesting that these spaces are progressively refined during training.
Comparing across language settings, we observe that alignment with concept spaces depends on language similarity and training data composition.

These findings have important implications for multilingual language model training, in particular in low-resource settings.
Since language-agnostic concept spaces emerge early and are relatively stable, multilingual training can focus on aligning languages to these spaces. Notably, our results indicate that relatively small amounts of high-quality, multi-aligned data improve alignment compared to \texttt{en}-only pivot data (phase one vs. phase two in EuroLLM). This suggests a promising path for closing the performance gap between high- and low-resource languages: improving alignment with concept spaces through targeted, multi-aligned data.

\section*{Acknowledgments}
FK and BP are supported by the ERC Consolidator Grant DIALECT 101043235. MME is supported by the Carlsberg Foundation, grant CF-25-0624. We are grateful to Dr. Mateusz Klimaszewski for giving us access to the EuroLLM checkpoints, and Dr. Kamil Deja for giving feedback on an earlier draft. We thank Pingjun Hong and Darja Jepifanova for providing their native speaker expertise, and the members of MaiNLP for many useful discussions.

\section*{Limitations}
In this work, we conduct a novel fine-grained investigation into how language-agnostic spaces emerge throughout multilingual LLM training. We mainly focus on only a single, relatively small (1.7B) model, yet provide additional supporting evidence on OLMo-2 7B and Apertus 8B (\cref{sec:other-models}). This choice is due to the limited availability of multilingual pretraining checkpoints at the time of writing. As noted in \cref{sec:related-work}, BLOOM \cite{workshop2023bloom176bparameteropenaccessmultilingual} was, until the recent release of Apertus \cite{DBLP:journals/corr/abs-2509-14233}, the only other state-of-the-art multilingual LM for which checkpoints are available, and offers only a very coarse granularity (4$-$8 checkpoints vs. our 26). We repeat experiments for target language pairs \texttt{xx--en} for pretraining checkpoints of Apertus 8B and OLMo-2 7B \cite{walsh2025}, a model trained on English data, with inadvertent multilingual ability, in \cref{sec:other-models}. These experiments suggest that our findings may hold for other multilingual models. However, further research, and in particular transparent and open multilingual pretraining is needed to understand the generalizability of our findings and the impact of model size and pretraining strategies on the development of shared concept spaces.

Furthermore, though we study a larger and more diverse set of concepts than previous work, we focus only on nouns. Further work is needed to understand how representations for other word classes behave and whether our findings apply.

Since we use Multi-SimLex as the pool for our concepts and their translations, we inherit its biases. In particular, the expected translation for a particular word may be misleading. For example, as discussed in \cref{sec:en-fr}, the translation for ``wife'' in French in the dataset is ``femme'', but this word also means ``woman''. However, we introduce a manual error analysis (\cref{sec:manual-eval}), capturing such biases. Because we also study lower-resourced languages, this analysis strengthens the evaluation compared to previous studies, which either relied on automatic machine translation \cite{wendler-etal-2024-llamas} or linguistic resources \cite{dumas-etal-2025-separating} to evaluate output. Such tools are often less reliable for lower-resourced languages \cite{Peppin2025TheMD}. Our approach therefore improves the evaluation methodology itself, allowing for a more equitable treatment of lower-resourced languages.

Finally, we focus only on the task of word-level translation. This is a fairly simple task, and language-agnostic spaces may form and behave differently for more complex tasks like sentence-level translation, open-ended generation, or reasoning. The focus on word-level translation is motivated by two factors. First, activation patching relies on a simple, controlled task (see discussion of confounding ``degrees of freedom'' in \citealp{DBLP:journals/corr/abs-2404-15255}). Second, we study pretraining checkpoints, specifically without any instruction tuning, which should improve the model's ability to follow more complex task instructions \cite{wei2022finetuned}. Therefore, focusing on this simple task reduces the confounding factor of whether the model can perform the task. This is confirmed by our preliminary experiments in \cref{sec:results-word-level}, which show that even early checkpoints can follow the task, in particular for high-resourced language pairs. We leave investigation of how language-agnostic spaces used for more complex multilingual tasks emerge and behave throughout multilingual pretraining to future work.

\section*{Ethical Considerations}
We acknowledge the use of ChatGPT for paraphrasing and lexical suggestions, with output checked carefully to avoid changes in meaning. In addition, ChatGPT and Github Copilot provided coding assistance.

\bibliography{anthology_small,clean,checked}

\begin{thebibliography}{40}
\providecommand{\natexlab}[1]{#1}

\bibitem[{Blevins and Zettlemoyer(2022)}]{blevins-zettlemoyer-2022-language}
Terra Blevins and Luke Zettlemoyer. 2022.
\newblock \href {https://doi.org/10.18653/v1/2022.emnlp-main.233} {Language contamination helps explains the cross-lingual capabilities of {E}nglish pretrained models}.
\newblock In \emph{Proceedings of the 2022 Conference on Empirical Methods in Natural Language Processing}, pages 3563--3574, Abu Dhabi, United Arab Emirates. Association for Computational Linguistics.

\bibitem[{Cao et~al.(2024)Cao, Chen, Jin, Chen, Liu, and Zhao}]{DBLP:journals/corr/abs-2411-17401}
Pengfei Cao, Yuheng Chen, Zhuoran Jin, Yubo Chen, Kang Liu, and Jun Zhao. 2024.
\newblock \href {https://doi.org/10.48550/arXiv.2411.17401} {One mind, many tongues: A deep dive into language-agnostic knowledge neurons in large language models}.
\newblock \emph{CoRR}, abs/2411.17401.

\bibitem[{Chen et~al.(2024)Chen, Cao, Chen, Liu, and Zhao}]{10.1609/aaai.v38i16.29735}
Yuheng Chen, Pengfei Cao, Yubo Chen, Kang Liu, and Jun Zhao. 2024.
\newblock \href {https://doi.org/10.1609/aaai.v38i16.29735} {Journey to the center of the knowledge neurons: discoveries of language-independent knowledge neurons and degenerate knowledge neurons}.
\newblock In \emph{Proceedings of the Thirty-Eighth AAAI Conference on Artificial Intelligence and Thirty-Sixth Conference on Innovative Applications of Artificial Intelligence and Fourteenth Symposium on Educational Advances in Artificial Intelligence}, AAAI'24/IAAI'24/EAAI'24. AAAI Press.

\bibitem[{Chirkova and Nikoulina(2024)}]{chirkova-nikoulina-2024-zero-shot}
Nadezhda Chirkova and Vassilina Nikoulina. 2024.
\newblock \href {https://doi.org/10.18653/v1/2024.inlg-main.53} {Zero-shot cross-lingual transfer in instruction tuning of large language models}.
\newblock In \emph{Proceedings of the 17th International Natural Language Generation Conference}, pages 695--708, Tokyo, Japan. Association for Computational Linguistics.

\bibitem[{Dumas et~al.(2025)Dumas, Wendler, Veselovsky, Monea, and West}]{dumas-etal-2025-separating}
Cl{\'e}ment Dumas, Chris Wendler, Veniamin Veselovsky, Giovanni Monea, and Robert West. 2025.
\newblock \href {https://doi.org/10.18653/v1/2025.acl-long.1536} {Separating tongue from thought: Activation patching reveals language-agnostic concept representations in transformers}.
\newblock In \emph{Proceedings of the 63rd Annual Meeting of the Association for Computational Linguistics (Volume 1: Long Papers)}, pages 31822--31841, Vienna, Austria. Association for Computational Linguistics.

\bibitem[{Feucht et~al.(2025)Feucht, Todd, Wallace, and Bau}]{feucht2025the}
Sheridan Feucht, Eric Todd, Byron~C Wallace, and David Bau. 2025.
\newblock \href {https://openreview.net/forum?id=bNTrKqqnG9} {The dual-route model of induction}.
\newblock In \emph{Second Conference on Language Modeling}.

\bibitem[{Grave et~al.(2018)Grave, Bojanowski, Gupta, Joulin, and Mikolov}]{grave-etal-2018-learning}
Edouard Grave, Piotr Bojanowski, Prakhar Gupta, Armand Joulin, and Tomas Mikolov. 2018.
\newblock \href {https://aclanthology.org/L18-1550/} {Learning word vectors for 157 languages}.
\newblock In \emph{Proceedings of the Eleventh International Conference on Language Resources and Evaluation ({LREC} 2018)}, Miyazaki, Japan. European Language Resources Association (ELRA).

\bibitem[{Heimersheim and Nanda(2024)}]{DBLP:journals/corr/abs-2404-15255}
Stefan Heimersheim and Neel Nanda. 2024.
\newblock \href {https://doi.org/10.48550/arXiv.2404.15255} {How to use and interpret activation patching}.
\newblock \emph{CoRR}, abs/2404.15255.

\bibitem[{Hernández-Cano et~al.(2025)Hernández-Cano, Hägele, Huang, Romanou, i~Llaquet, Pásztor, Messmer, Garbaya, Durech, Hakimi, Giraldo, Ismayilzada, Foroutan, Moalla, Chen, Sabolcec, Xu, Aerni, AlKhamissi, Marinas, Amani, Ansaripour, Badanin, Benoit, Boros, Browning, Bösch, Böther, Canova, Challier, Charmillot, Coles, Deriu, Devos, Drescher, Dzenhaliou, Ehrmann, Fan, Fan, Gao, Gila, Grandury, Hashemi, Hoyle, Jiang, Klein, Kucharavy, Kucherenko, Lübeck, Machacek, Manitaras, Marfurt, Matoba, Matrenok, Mendonça, Mohamed, Montariol, Mouchel, Najem-Meyer, Ni, Oliva, Pagliardini, Palme, Panferov, Paoletti, Passerini, Pavlov, Poiroux, Ponkshe, Ranchin, Rando, Sauser, Saydaliev, Sayfiddinov, Schneider, Schuppli, Scialanga, Semenov, Shridhar, Singhal, Sotnikova, Sternfeld, Tarun, Teiletche, Vamvas, Yao, Ilic, Klimovic, Krause, Gulcehre, Rosenthal, Ash, Tramèr, VandeVondele, Veraldi, Rajman, Schulthess, Hoefler, Bosselut, Jaggi, and Schlag}]{DBLP:journals/corr/abs-2509-14233}
Alejandro Hernández-Cano, Alexander Hägele, Allen~Hao Huang, Angelika Romanou, Antoni-Joan~Solergibert i~Llaquet, Barna Pásztor, Bettina Messmer, Dhia Garbaya, Eduard~Frank Durech, Ido Hakimi, Juan~García Giraldo, Mete Ismayilzada, Negar Foroutan, Skander Moalla, Tiancheng Chen, Vinko Sabolcec, Yixuan Xu, Michael Aerni, Badr AlKhamissi, and 82 others. 2025.
\newblock \href {https://doi.org/10.48550/arXiv.2509.14233} {Apertus: Democratizing open and compliant llms for global language environments}.
\newblock \emph{CoRR}, abs/2509.14233.

\bibitem[{Joshi et~al.(2020)Joshi, Santy, Budhiraja, Bali, and Choudhury}]{joshi-etal-2020-state}
Pratik Joshi, Sebastin Santy, Amar Budhiraja, Kalika Bali, and Monojit Choudhury. 2020.
\newblock \href {https://doi.org/10.18653/v1/2020.acl-main.560} {The state and fate of linguistic diversity and inclusion in the {NLP} world}.
\newblock In \emph{Proceedings of the 58th Annual Meeting of the Association for Computational Linguistics}, pages 6282--6293, Online. Association for Computational Linguistics.

\bibitem[{Kojima et~al.(2024)Kojima, Okimura, Iwasawa, Yanaka, and Matsuo}]{kojima-etal-2024-multilingual}
Takeshi Kojima, Itsuki Okimura, Yusuke Iwasawa, Hitomi Yanaka, and Yutaka Matsuo. 2024.
\newblock \href {https://doi.org/10.18653/v1/2024.naacl-long.384} {On the multilingual ability of decoder-based pre-trained language models: Finding and controlling language-specific neurons}.
\newblock In \emph{Proceedings of the 2024 Conference of the North American Chapter of the Association for Computational Linguistics: Human Language Technologies (Volume 1: Long Papers)}, pages 6919--6971, Mexico City, Mexico. Association for Computational Linguistics.

\bibitem[{Lee et~al.(2025)Lee, Woo, Ko, and Son}]{lee-etal-2025-controlling}
Nahyun Lee, Yeongseo Woo, Hyunwoo Ko, and Guijin Son. 2025.
\newblock \href {https://doi.org/10.18653/v1/2025.acl-srw.81} {Controlling language confusion in multilingual {LLM}s}.
\newblock In \emph{Proceedings of the 63rd Annual Meeting of the Association for Computational Linguistics (Volume 4: Student Research Workshop)}, pages 1026--1035, Vienna, Austria. Association for Computational Linguistics.

\bibitem[{Li et~al.(2024)Li, Huang, Ching, Dai, and Chen}]{li-etal-2024-prealign}
Jiahuan Li, Shujian Huang, Aarron Ching, Xinyu Dai, and Jiajun Chen. 2024.
\newblock \href {https://doi.org/10.18653/v1/2024.emnlp-main.572} {{P}re{A}lign: Boosting cross-lingual transfer by early establishment of multilingual alignment}.
\newblock In \emph{Proceedings of the 2024 Conference on Empirical Methods in Natural Language Processing}, pages 10246--10257, Miami, Florida, USA. Association for Computational Linguistics.

\bibitem[{Lin et~al.(2025)Lin, Martins, and Schuetze}]{lin-etal-2025-recipe}
Peiqin Lin, Andre Martins, and Hinrich Schuetze. 2025.
\newblock \href {https://doi.org/10.18653/v1/2025.findings-naacl.225} {A recipe of parallel corpora exploitation for multilingual large language models}.
\newblock In \emph{Findings of the Association for Computational Linguistics: NAACL 2025}, pages 4038--4050, Albuquerque, New Mexico. Association for Computational Linguistics.

\bibitem[{Liu and Fu(2024)}]{Liu2024ResponsibleMLA}
Junhua Liu and Bin Fu. 2024.
\newblock \href {https://arxiv.org/abs/2410.17532v1} {Responsible multilingual large language models: A survey of development, applications, and societal impact}.
\newblock \emph{arXiv:2410.17532v1}.

\bibitem[{Martins et~al.(2025)Martins, Fernandes, Alves, Guerreiro, Rei, Alves, Pombal, Farajian, Faysse, Klimaszewski, Colombo, Haddow, de~Souza, Birch, and Martins}]{martins2024eurollmmultilinguallanguagemodels}
Pedro~Henrique Martins, Patrick Fernandes, João Alves, Nuno~M. Guerreiro, Ricardo Rei, Duarte~M. Alves, José Pombal, Amin Farajian, Manuel Faysse, Mateusz Klimaszewski, Pierre Colombo, Barry Haddow, José G.~C. de~Souza, Alexandra Birch, and André F.~T. Martins. 2025.
\newblock \href {https://doi.org/10.1016/j.procs.2025.02.260} {Eurollm: Multilingual language models for europe}.
\newblock \emph{Procedia Computer Science}.

\bibitem[{Mueller et~al.(2024)Mueller, Brinkmann, Li, Marks, Pal, Prakash, Rager, Sankaranarayanan, Sharma, Sun, Todd, Bau, and Belinkov}]{DBLP:journals/corr/abs-2408-01416}
Aaron Mueller, Jannik Brinkmann, Millicent~L. Li, Samuel Marks, Koyena Pal, Nikhil Prakash, Can Rager, Aruna Sankaranarayanan, Arnab~Sen Sharma, Jiuding Sun, Eric Todd, David Bau, and Yonatan Belinkov. 2024.
\newblock \href {https://doi.org/10.48550/arXiv.2408.01416} {The quest for the right mediator: A history, survey, and theoretical grounding of causal interpretability}.
\newblock \emph{CoRR}, abs/2408.01416.

\bibitem[{Navigli and Ponzetto(2010)}]{navigli-ponzetto-2010-babelnet}
Roberto Navigli and Simone~Paolo Ponzetto. 2010.
\newblock \href {https://aclanthology.org/P10-1023/} {{B}abel{N}et: Building a very large multilingual semantic network}.
\newblock In \emph{Proceedings of the 48th Annual Meeting of the Association for Computational Linguistics}, pages 216--225, Uppsala, Sweden. Association for Computational Linguistics.

\bibitem[{Peng and S{\o}gaard(2024)}]{peng-sogaard-2024-concept}
Qiwei Peng and Anders S{\o}gaard. 2024.
\newblock \href {https://doi.org/10.18653/v1/2024.emnlp-main.315} {Concept space alignment in multilingual {LLM}s}.
\newblock In \emph{Proceedings of the 2024 Conference on Empirical Methods in Natural Language Processing}, pages 5511--5526, Miami, Florida, USA. Association for Computational Linguistics.

\bibitem[{Peppin et~al.(2025)Peppin, Kreutzer, Sebag, Marchisio, Ermis, Dang, Cahyawijaya, Singh, Goldfarb-Tarrant, Aryabumi, Aakanksha, Ko, Üstün, Gallé, Fadaee, and Hooker}]{Peppin2025TheMD}
Aidan Peppin, Julia Kreutzer, Alice~Schoenauer Sebag, Kelly Marchisio, Beyza Ermis, John Dang, Samuel Cahyawijaya, Shivalika Singh, Seraphina Goldfarb-Tarrant, Viraat Aryabumi, Aakanksha, Wei-Yin Ko, Ahmet Üstün, Matthias Gallé, Marzieh Fadaee, and Sara Hooker. 2025.
\newblock \href {https://arxiv.org/abs/2505.21344v1} {The multilingual divide and its impact on global ai safety}.
\newblock \emph{arXiv:2505.21344v1}.

\bibitem[{Ranathunga and de~Silva(2022)}]{ranathunga-de-silva-2022-languages}
Surangika Ranathunga and Nisansa de~Silva. 2022.
\newblock \href {https://doi.org/10.18653/v1/2022.aacl-main.62} {Some languages are more equal than others: Probing deeper into the linguistic disparity in the {NLP} world}.
\newblock In \emph{Proceedings of the 2nd Conference of the Asia-Pacific Chapter of the Association for Computational Linguistics and the 12th International Joint Conference on Natural Language Processing (Volume 1: Long Papers)}, pages 823--848, Online only. Association for Computational Linguistics.

\bibitem[{Riemenschneider and Frank(2025)}]{riemenschneider-frank-2025-cross}
Frederick Riemenschneider and Anette Frank. 2025.
\newblock \href {https://doi.org/10.18653/v1/2025.acl-long.661} {Cross-lingual generalization and compression: From language-specific to shared neurons}.
\newblock In \emph{Proceedings of the 63rd Annual Meeting of the Association for Computational Linguistics (Volume 1: Long Papers)}, pages 13470--13491, Vienna, Austria. Association for Computational Linguistics.

\bibitem[{Schut et~al.(2025)Schut, Gal, and Farquhar}]{schut2025do}
Lisa Schut, Yarin Gal, and Sebastian Farquhar. 2025.
\newblock \href {https://openreview.net/forum?id=I8BOtOPcOv} {Do multilingual {LLM}s think in english?}
\newblock In \emph{ICLR 2025 Workshop on Building Trust in Language Models and Applications}.

\bibitem[{Shen et~al.(2025)Shen, Lai, Wang, Gao, Luo, Fraser, and Sun}]{shen-etal-2025-unaligned}
Yingli Shen, Wen Lai, Shuo Wang, Ge~Gao, Kangyang Luo, Alexander Fraser, and Maosong Sun. 2025.
\newblock \href {https://doi.org/10.18653/v1/2025.emnlp-main.374} {From unaligned to aligned: Scaling multilingual {LLM}s with multi-way parallel corpora}.
\newblock In \emph{Proceedings of the 2025 Conference on Empirical Methods in Natural Language Processing}, pages 7357--7379, Suzhou, China. Association for Computational Linguistics.

\bibitem[{Soldaini et~al.(2024)Soldaini, Kinney, Bhagia, Schwenk, Atkinson, Authur, Bogin, Chandu, Dumas, Elazar, Hofmann, Jha, Kumar, Lucy, Lyu, Lambert, Magnusson, Morrison, Muennighoff, Naik, Nam, Peters, Ravichander, Richardson, Shen, Strubell, Subramani, Tafjord, Walsh, Zettlemoyer, Smith, Hajishirzi, Beltagy, Groeneveld, Dodge, and Lo}]{soldaini-etal-2024-dolma}
Luca Soldaini, Rodney Kinney, Akshita Bhagia, Dustin Schwenk, David Atkinson, Russell Authur, Ben Bogin, Khyathi Chandu, Jennifer Dumas, Yanai Elazar, Valentin Hofmann, Ananya Jha, Sachin Kumar, Li~Lucy, Xinxi Lyu, Nathan Lambert, Ian Magnusson, Jacob Morrison, Niklas Muennighoff, and 17 others. 2024.
\newblock \href {https://doi.org/10.18653/v1/2024.acl-long.840} {Dolma: an open corpus of three trillion tokens for language model pretraining research}.
\newblock In \emph{Proceedings of the 62nd Annual Meeting of the Association for Computational Linguistics (Volume 1: Long Papers)}, pages 15725--15788, Bangkok, Thailand. Association for Computational Linguistics.

\bibitem[{Stanczak et~al.(2022)Stanczak, Ponti, Torroba~Hennigen, Cotterell, and Augenstein}]{stanczak-etal-2022-neurons}
Karolina Stanczak, Edoardo Ponti, Lucas Torroba~Hennigen, Ryan Cotterell, and Isabelle Augenstein. 2022.
\newblock \href {https://doi.org/10.18653/v1/2022.naacl-main.114} {Same neurons, different languages: Probing morphosyntax in multilingual pre-trained models}.
\newblock In \emph{Proceedings of the 2022 Conference of the North American Chapter of the Association for Computational Linguistics: Human Language Technologies}, pages 1589--1598, Seattle, United States. Association for Computational Linguistics.

\bibitem[{Tang et~al.(2024)Tang, Luo, Huang, Zhang, Wang, Zhao, Wei, and Wen}]{tang-etal-2024-language}
Tianyi Tang, Wenyang Luo, Haoyang Huang, Dongdong Zhang, Xiaolei Wang, Xin Zhao, Furu Wei, and Ji-Rong Wen. 2024.
\newblock \href {https://doi.org/10.18653/v1/2024.acl-long.309} {Language-specific neurons: The key to multilingual capabilities in large language models}.
\newblock In \emph{Proceedings of the 62nd Annual Meeting of the Association for Computational Linguistics (Volume 1: Long Papers)}, pages 5701--5715, Bangkok, Thailand. Association for Computational Linguistics.

\bibitem[{Tezuka and Inoue(2025)}]{tezuka-inoue-2025-transfer}
Hinata Tezuka and Naoya Inoue. 2025.
\newblock \href {https://doi.org/10.18653/v1/2025.emnlp-main.1618} {The transfer neurons hypothesis: An underlying mechanism for language latent space transitions in multilingual {LLM}s}.
\newblock In \emph{Proceedings of the 2025 Conference on Empirical Methods in Natural Language Processing}, pages 31742--31792, Suzhou, China. Association for Computational Linguistics.

\bibitem[{Vig et~al.(2020)Vig, Gehrmann, Belinkov, Qian, Nevo, Singer, and Shieber}]{10.5555/3495724.3496763}
Jesse Vig, Sebastian Gehrmann, Yonatan Belinkov, Sharon Qian, Daniel Nevo, Yaron Singer, and Stuart Shieber. 2020.
\newblock Investigating gender bias in language models using causal mediation analysis.
\newblock In \emph{Proceedings of the 34th International Conference on Neural Information Processing Systems}, NIPS '20, Red Hook, NY, USA. Curran Associates Inc.

\bibitem[{Vuli{\'c} et~al.(2020)Vuli{\'c}, Baker, Ponti, Petti, Leviant, Wing, Majewska, Bar, Malone, Poibeau, Reichart, and Korhonen}]{vulic-etal-2020-multi}
Ivan Vuli{\'c}, Simon Baker, Edoardo~Maria Ponti, Ulla Petti, Ira Leviant, Kelly Wing, Olga Majewska, Eden Bar, Matt Malone, Thierry Poibeau, Roi Reichart, and Anna Korhonen. 2020.
\newblock \href {https://doi.org/10.1162/coli_a_00391} {Multi-{S}im{L}ex: A large-scale evaluation of multilingual and crosslingual lexical semantic similarity}.
\newblock \emph{Computational Linguistics}, 46(4):847--897.

\bibitem[{Walsh et~al.(2025)Walsh, Soldaini, Groeneveld, Lo, Arora, Bhagia, Gu, Huang, Jordan, Lambert, Schwenk, Tafjord, Anderson, Atkinson, Brahman, Clark, Dasigi, Dziri, Ettinger, Guerquin, Heineman, Ivison, Koh, Liu, Malik, Merrill, Miranda, Morrison, Murray, Nam, Poznanski, Pyatkin, Rangapur, Schmitz, Skjonsberg, Wadden, Wilhelm, Wilson, Zettlemoyer, Farhadi, Smith, and Hajishirzi}]{walsh2025}
Evan~Pete Walsh, Luca Soldaini, Dirk Groeneveld, Kyle Lo, Shane Arora, Akshita Bhagia, Yuling Gu, Shengyi Huang, Matt Jordan, Nathan Lambert, Dustin Schwenk, Oyvind Tafjord, Taira Anderson, David Atkinson, Faeze Brahman, Christopher Clark, Pradeep Dasigi, Nouha Dziri, Allyson Ettinger, and 23 others. 2025.
\newblock \href {https://openreview.net/forum?id=2ezugTT9kU} {2 {OLM}o 2 furious ({COLM}{\textquoteright}s version)}.
\newblock In \emph{Second Conference on Language Modeling}.

\bibitem[{Wang et~al.(2024{\natexlab{a}})Wang, Minervini, and Ponti}]{wang-etal-2024-probing-emergence}
Hetong Wang, Pasquale Minervini, and Edoardo Ponti. 2024{\natexlab{a}}.
\newblock \href {https://doi.org/10.18653/v1/2024.findings-acl.724} {Probing the emergence of cross-lingual alignment during {LLM} training}.
\newblock In \emph{Findings of the Association for Computational Linguistics: ACL 2024}, pages 12159--12173, Bangkok, Thailand. Association for Computational Linguistics.

\bibitem[{Wang et~al.(2024{\natexlab{b}})Wang, Haddow, Peng, and Birch}]{DBLP:journals/corr/abs-2406-09265}
Weixuan Wang, Barry Haddow, Wei Peng, and Alexandra Birch. 2024{\natexlab{b}}.
\newblock \href {https://doi.org/10.48550/arXiv.2406.09265} {Sharing matters: Analysing neurons across languages and tasks in llms}.
\newblock \emph{CoRR}, abs/2406.09265.

\bibitem[{Wei et~al.(2022)Wei, Bosma, Zhao, Guu, Yu, Lester, Du, Dai, and Le}]{wei2022finetuned}
Jason Wei, Maarten Bosma, Vincent Zhao, Kelvin Guu, Adams~Wei Yu, Brian Lester, Nan Du, Andrew~M. Dai, and Quoc~V Le. 2022.
\newblock \href {https://openreview.net/forum?id=gEZrGCozdqR} {Finetuned language models are zero-shot learners}.
\newblock In \emph{International Conference on Learning Representations}.

\bibitem[{Wendler et~al.(2024)Wendler, Veselovsky, Monea, and West}]{wendler-etal-2024-llamas}
Chris Wendler, Veniamin Veselovsky, Giovanni Monea, and Robert West. 2024.
\newblock \href {https://doi.org/10.18653/v1/2024.acl-long.820} {Do llamas work in {E}nglish? on the latent language of multilingual transformers}.
\newblock In \emph{Proceedings of the 62nd Annual Meeting of the Association for Computational Linguistics (Volume 1: Long Papers)}, pages 15366--15394, Bangkok, Thailand. Association for Computational Linguistics.

\bibitem[{Workshop et~al.(2022)Workshop, :, Scao, Fan, Akiki, Pavlick, Ilić, Hesslow, Castagné, Luccioni, Yvon, Gallé, Tow, Rush, Biderman, Webson, Ammanamanchi, Wang, Sagot, Muennighoff, del Moral, Ruwase, Bawden, Bekman, McMillan-Major, Beltagy, Nguyen, Saulnier, Tan, Suarez, Sanh, Laurençon, Jernite, Launay, Mitchell, Raffel, Gokaslan, Simhi, Soroa, Aji, Alfassy, Rogers, Nitzav, Xu, Mou, Emezue, Klamm, Leong, van Strien, Adelani, Radev, Ponferrada, Levkovizh, Kim, Natan, Toni, Dupont, Kruszewski, Pistilli, Elsahar, Benyamina, Tran, Yu, Abdulmumin, Johnson, Gonzalez-Dios, de~la Rosa, Chim, Dodge, Zhu, Chang, Frohberg, Tobing, Bhattacharjee, Almubarak, Chen, Lo, Werra, Weber, Phan, allal, Tanguy, Dey, Muñoz, Masoud, Grandury, Šaško, Huang, Coavoux, Singh, Jiang, Vu, Jauhar, Ghaleb, Subramani, Kassner, Khamis, Nguyen, Espejel, de~Gibert, Villegas, Henderson, Colombo, Amuok, Lhoest, Harliman, Bommasani, López, Ribeiro, Osei, Pyysalo, Nagel, Bose, Muhammad, Sharma, Longpre, Nikpoor, Silberberg, Pai,
  Zink, Torrent, Schick, Thrush, Danchev, Nikoulina, Laippala, Lepercq, Prabhu, Alyafeai, Talat, Raja, Heinzerling, Si, Taşar, Salesky, Mielke, Lee, Sharma, Santilli, Chaffin, Stiegler, Datta, Szczechla, Chhablani, Wang, Pandey, Strobelt, Fries, Rozen, Gao, Sutawika, Bari, Al-shaibani, Manica, Nayak, Teehan, Albanie, Shen, Ben-David, Bach, Kim, Bers, Fevry, Neeraj, Thakker, Raunak, Tang, Yong, Sun, Brody, Uri, Tojarieh, Roberts, Chung, Tae, Phang, Press, Li, Narayanan, Bourfoune, Casper, Rasley, Ryabinin, Mishra, Zhang, Shoeybi, Peyrounette, Patry, Tazi, Sanseviero, von Platen, Cornette, Lavallée, Lacroix, Rajbhandari, Gandhi, Smith, Requena, Patil, Dettmers, Baruwa, Singh, Cheveleva, Ligozat, Subramonian, Névéol, Lovering, Garrette, Tunuguntla, Reiter, Taktasheva, Voloshina, Bogdanov, Winata, Schoelkopf, Kalo, Novikova, Forde, Clive, Kasai, Kawamura, Hazan, Carpuat, Clinciu, Kim, Cheng, Serikov, Antverg, van~der Wal, Zhang, Zhang, Gehrmann, Mirkin, Pais, Shavrina, Scialom, Yun, Limisiewicz, Rieser,
  Protasov, Mikhailov, Pruksachatkun, Belinkov, Bamberger, Kasner, Rueda, Pestana, Feizpour, Khan, Faranak, Santos, Hevia, Unldreaj, Aghagol, Abdollahi, Tammour, HajiHosseini, Behroozi, Ajibade, Saxena, Ferrandis, McDuff, Contractor, Lansky, David, Kiela, Nguyen, Tan, Baylor, Ozoani, Mirza, Ononiwu, Rezanejad, Jones, Bhattacharya, Solaiman, Sedenko, Nejadgholi, Passmore, Seltzer, Sanz, Dutra, Samagaio, Elbadri, Mieskes, Gerchick, Akinlolu, McKenna, Qiu, Ghauri, Burynok, Abrar, Rajani, Elkott, Fahmy, Samuel, An, Kromann, Hao, Alizadeh, Shubber, Wang, Roy, Viguier, Le, Oyebade, Le, Yang, Nguyen, Kashyap, Palasciano, Callahan, Shukla, Miranda-Escalada, Singh, Beilharz, Wang, Brito, Zhou, Jain, Xu, Fourrier, Periñán, Molano, Yu, Manjavacas, Barth, Fuhrimann, Altay, Bayrak, Burns, Vrabec, Bello, Dash, Kang, Giorgi, Golde, Posada, Sivaraman, Bulchandani, Liu, Shinzato, de~Bykhovetz, Takeuchi, Pàmies, Castillo, Nezhurina, Sänger, Samwald, Cullan, Weinberg, Wolf, Mihaljcic, Liu, Freidank, Kang, Seelam, Dahlberg,
  Broad, Muellner, Fung, Haller, Chandrasekhar, Eisenberg, Martin, Canalli, Su, Su, Cahyawijaya, Garda, Deshmukh, Mishra, Kiblawi, Ott, Sang-aroonsiri, Kumar, Schweter, Bharati, Laud, Gigant, Kainuma, Kusa, Labrak, Bajaj, Venkatraman, Xu, Xu, Xu, Tan, Xie, Ye, Bras, Belkada, and Wolf}]{workshop2023bloom176bparameteropenaccessmultilingual}
BigScience Workshop, :, Teven~Le Scao, Angela Fan, Christopher Akiki, Ellie Pavlick, Suzana Ilić, Daniel Hesslow, Roman Castagné, Alexandra~Sasha Luccioni, François Yvon, Matthias Gallé, Jonathan Tow, Alexander~M. Rush, Stella Biderman, Albert Webson, Pawan~Sasanka Ammanamanchi, Thomas Wang, Benoît Sagot, and 375 others. 2022.
\newblock \href {https://arxiv.org/abs/2211.05100v4} {Bloom: A 176b-parameter open-access multilingual language model}.
\newblock \emph{arXiv:2211.05100v4}.

\bibitem[{Zeng et~al.(2025)Zeng, Han, Chen, and Yu}]{zeng-etal-2025-converging}
Hongchuan Zeng, Senyu Han, Lu~Chen, and Kai Yu. 2025.
\newblock \href {https://aclanthology.org/2025.coling-main.707/} {Converging to a lingua franca: Evolution of linguistic regions and semantics alignment in multilingual large language models}.
\newblock In \emph{Proceedings of the 31st International Conference on Computational Linguistics}, pages 10602--10617, Abu Dhabi, UAE. Association for Computational Linguistics.

\bibitem[{Zhang et~al.(2025)Zhang, Lai, Liu, She, Liu, Gong, Huang, and Chen}]{Zhang2025HowDA}
Shimao Zhang, Zhejian Lai, Xiang Liu, Shuaijie She, Xiao Liu, Yeyun Gong, Shujian Huang, and Jiajun Chen. 2025.
\newblock \href {https://api.semanticscholar.org/CorpusID:278911959} {How does alignment enhance llms' multilingual capabilities? a language neurons perspective}.
\newblock \emph{ArXiv}, abs/2505.21505.

\bibitem[{Zhao et~al.(2024)Zhao, Zhang, Chen, Kawaguchi, and Bing}]{NEURIPS2024_1bd359b3}
Yiran Zhao, Wenxuan Zhang, Guizhen Chen, Kenji Kawaguchi, and Lidong Bing. 2024.
\newblock \href {https://doi.org/10.52202/079017-0489} {How do large language models handle multilingualism?}
\newblock In \emph{Advances in Neural Information Processing Systems}, volume~37, pages 15296--15319. Curran Associates, Inc.

\bibitem[{Zhong et~al.(2025)Zhong, Liu, Cheng, Jiang, Wan, Chu, Murawaki, and Kurohashi}]{zhong-etal-2025-language}
Chengzhi Zhong, Qianying Liu, Fei Cheng, Junfeng Jiang, Zhen Wan, Chenhui Chu, Yugo Murawaki, and Sadao Kurohashi. 2025.
\newblock \href {https://doi.org/10.18653/v1/2025.findings-acl.1350} {What language do non-{E}nglish-centric large language models think in?}
\newblock In \emph{Findings of the Association for Computational Linguistics: ACL 2025}, pages 26333--26346, Vienna, Austria. Association for Computational Linguistics.

\end{thebibliography}
\appendix
\crefalias{section}{appendix}
\section{Appendix}
\subsection{Technical Details}\label{sec:tech}
We build on the implementation of concept patching published by \citealp{dumas-etal-2025-separating}\footnote{\url{https://github.com/Butanium/llm-lang-agnostic}}. Specifically, we modify the prompt construction to produce concept-aligned prompts for both source and target prompt sets across all tested language pairs. We also adapt it to prevent overlap between few-shot examples on the source and target side. Furthermore, we update it to use nnsight 0.5.5\footnote{\url{https://nnsight.net}} in order to enable multi-token generation under patching.

We generate five next tokens, and parse output up to the first generated quotation mark. If an initial substring matched the target or a synonym, we count this as correct, to avoid penalizing overly long continuations (for example: ``absence of desire'', when the expected word is ``absence'').

Please refer to the implementation at \url{https://github.com/mainlp/shared-concept-spaces} for more information.

\paragraph{Choice of Layer}
\citealp{dumas-etal-2025-separating} find that the choice of layer to patch from doesn't significantly affect whether patching can induce the source concept, so long as it is not one of the later layers. We conduct initial experiments for target language pairs \texttt{xx--en} to confirm this, and find that we can patch up to about layer~14, after which the concept can no longer be reliably induced. \cref{tab:layer-corr} shows the Pearson correlation coefficient between the tested layers (6, 8, 9, 11, 12, 14, 16) and layer 10 computed over the full delta accuracy curves for \texttt{seen} over unpatched across checkpoints. A high correlation (>0.9) indicates that the trajectory of the patching effect over checkpoints is consistent between layers.

\begin{table}[t]
\centering
\small
\begin{tabular}{l@{\hskip 8pt}r@{\hskip 8pt}r@{\hskip 8pt}r@{\hskip 8pt}r@{\hskip 8pt}r@{\hskip 8pt}r@{\hskip 8pt}r}
\toprule
tgt pair & L6 & L8 & L9 & L11 & L12 & L14 & L16 \\
\midrule
\texttt{es--en} & 0.93 & 0.96 & 0.97 & 0.97 & 0.95 & 0.92 & 0.08 \\
\texttt{fr--en} & 0.97 & 0.99 & 0.99 & 0.99 & 0.98 & 0.95 & -0.06 \\
\texttt{zh--en} & 0.93 & 0.95 & 0.98 & 0.99 & 0.97 & 0.93 & 0.00 \\
\texttt{pl--en} & 0.95 & 0.98 & 0.98 & 0.99 & 0.97 & 0.94 & 0.63 \\
\texttt{ru--en} & 0.99 & 0.99 & 1.00 & 1.00 & 0.99 & 0.97 & 0.74 \\
\texttt{et--en} & 0.99 & 0.99 & 0.99 & 1.00 & 0.99 & 0.97 & 0.67 \\
\texttt{fi--en} & 0.99 & 1.00 &  1.00 & 1.00 & 1.00 & 0.99 & 0.68 \\
\texttt{sw--en} & 0.99 & 1.00 & 1.00 & 1.00 & 1.00 & 1.00 & 0.91 \\
\bottomrule
\end{tabular}
\caption{Pearson correlation of the difference in translation accuracy curves (\texttt{seen} vs.~unpatched) over checkpoints between each tested layer and layer~10, for target language pairs \texttt{xx--en}. A high correlation (>0.9) indicates a consistent trend between checkpoints.}
\label{tab:layer-corr}
\end{table}

\subsection{Synonym Expansion}\label{sec:babelnet}
We attempt to follow prior work in expanding our translation dataset with synonyms from BabelNet \cite{navigli-ponzetto-2010-babelnet}, which would capture more correct translations of a particular concept. However, we find the resulting expansions to be overly generous, even after filtering by quality tags (as provided by BabelNet) and for nouns. This is likely partially due to the nature of our concept set, which includes abstract and infrequent terms (e.g., ``afterworld'', ``acetylcholine''). For example, ``acid'' is expanded to many different street names for the drug ``LSD''. Such expansions are difficult to filter out automatically, especially considering our treatment of low-resourced languages, such as Welsh and Swahili. For such languages, linguistic resources and tools are typically of lesser quality \cite{Peppin2025TheMD}. Furthermore, too generous expansions across 11 tested languages results in very few compatible source and target concept pairs where a compatible pair has no overlapping words across all languages. Therefore, we instead adapt our evaluation as described in \cref{sec:manual-eval}, manually inspecting model outputs to gain a more accurate understanding of the effect induced by cross-lingual concept patching.

\clearpage
\subsection{Example Prompts}\label{sec:example-prompts}
Here, we provide some example prompts for the different settings. For target language pair \texttt{fr--en}, a full target prompt for the target concept ``triumph'' is shown below.
\begin{mdframed}[backgroundcolor=blue!5, linecolor=blue, linewidth=0.5pt]
Français: ``août'' - English: ``august''\\
Français: ``animal'' - English: ``animal''\\
Français: ``critique'' - English: ``criticism''\\
Français: ``base'' - English: ``base''\\
Français: ``cage'' - English: ``cage''\\
Français: ``triomphe'' - English: ``
\end{mdframed}
In the following sections, we show examples of source prompts for the source concept ``curtain'' in different settings, given this target prompt.

\paragraph{unpatched Setting}
As a baseline, we run inference on the target prompt \textit{without} any intervention. This prompt is exactly the same as the one used in \texttt{tgt} setting, the difference is that here, the prompt is run without intervention.

\begin{mdframed}[backgroundcolor=gray!5, linecolor=gray, linewidth=0.5pt]
Français: ``pollution'' - English: ``pollution''\\
Français: ``dépression'' - English: ``depression''\\
Français: ``structure'' - English: ``structure''\\
Français: ``action'' - English: ``action''\\
Français: ``os'' - English: ``bone''\\
Français: ``rideau'' - English: ``
\end{mdframed}

\paragraph{\texttt{tgt} Setting Source Prompts}
In this ablation setting, the source and target language pairs are the same, the prompts differ only in the few-shot examples and the concept to be translated.
\begin{mdframed}[backgroundcolor=red!5, linecolor=red, linewidth=0.5pt]
Français: ``pollution'' - English: ``pollution''\\
Français: ``dépression'' - English: ``depression''\\
Français: ``structure'' - English: ``structure''\\
Français: ``action'' - English: ``action''\\
Français: ``os'' - English: ``bone''\\
Français: ``rideau'' - English: ``
\end{mdframed}

\paragraph{\texttt{en\_en} Setting Source Prompts}
In this control setting, the model must \textit{copy} the English word, rather than translate it.
\begin{mdframed}[backgroundcolor=red!5, linecolor=red, linewidth=0.5pt]
English: ``pollution'' - English: ``pollution''\\
English: ``depression'' - English: ``depression''\\
English: ``structure'' - English: ``structure''\\
English: ``action'' - English: ``action''\\
English: ``bone'' - English: ``bone''\\
English: ``curtain'' - English: ``
\end{mdframed}

\paragraph{\texttt{seen} Setting Source Prompts}
The language pairs in \texttt{seen} are in EuroLLM's training data, \texttt{excluding} \texttt{en} and the languages in the target pair. For this example, \texttt{seen} includes all ordered pairs of \texttt{es}, \texttt{zh}, \texttt{pl}, \texttt{ru}, \texttt{et}, \texttt{fi}, \texttt{yue}, for 42 total source language pairs. Below, we show a few sample prompts.

\begin{mdframed}[backgroundcolor=red!5, linecolor=red, linewidth=0.5pt]
Suomi: ``saastuminen'' - Polski: ``zanieczyszczenie''\\
Suomi: ``masennus'' - Polski: ``depresja''\\
Suomi: ``rakenne'' - Polski: ``struktura''\\
Suomi: ``toiminta'' - Polski: ``akcja''\\
Suomi: ``luu'' - Polski: ``kość''\\
Suomi: ``verho'' - Polski: ``
\end{mdframed}

\begin{mdframed}[backgroundcolor=red!5, linecolor=red, linewidth=0.5pt]
Polski: ``zanieczyszczenie'' - Español: ``contaminación''\\
Polski: ``depresja'' - Español: ``depresión''\\
Polski: ``struktura'' - Español: ``estructura''\\
Polski: ``akcja'' - Español: ``acción''\\
Polski: ``kość'' - Español: ``hueso''\\
Polski: ``zasłona'' - Español: ``
\end{mdframed}

\begin{mdframed}[backgroundcolor=red!5, linecolor=red, linewidth=0.5pt]
\foreignlanguage{russian}{Русский: ``загрязнение''} - Suomi: ``saastuminen''\\
\foreignlanguage{russian}{Русский: ``депрессия''} - Suomi: ``masennus''\\
\foreignlanguage{russian}{Русский: ``структура''} - Suomi: ``rakenne''\\
\foreignlanguage{russian}{Русский: ``действие''} - Suomi: ``toiminta''\\
\foreignlanguage{russian}{Русский: ``кость''} - Suomi: ``luu''\\
\foreignlanguage{russian}{Русский: ``занавеска''} - Suomi: ``
\end{mdframed}

\subsection{Manual Annotation}\label{sec:annotation}
\begin{figure*}
\centering
\includegraphics[height=0.9\textheight, keepaspectratio]{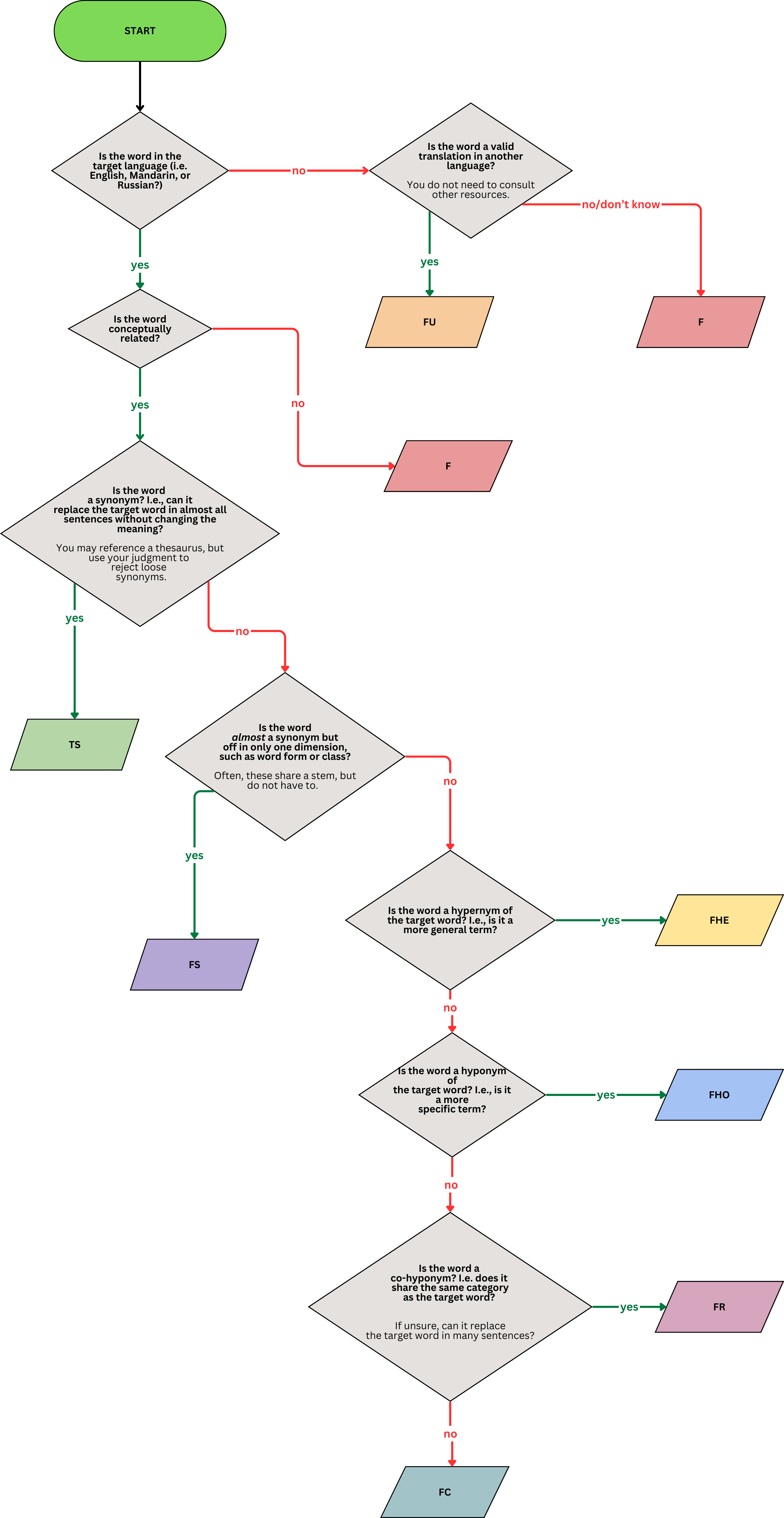}
\caption{The decision tree used to annotate errors as described in \cref{sec:manual-eval}. \texttt{T} is not included, as this was tagged automatically.}
\label{fig:decision-tree}
\end{figure*}
The in-house annotators (one per output language) are native speakers of the respective languages. All three annotators label a small subset of the English outputs (120 examples total). We exclude \texttt{T} from this set as it is automatically tagged, and stratify over the remaining classes, except for \texttt{F}. \texttt{F} is the majority class and typically most straightforward to tag, as many outputs, in particular in the very early stages of training (first two steps), are gibberish, or immediately recognizable as unrelated. Therefore, we down-sample \texttt{F} to 20/120. We compute a Fleiss’ Kappa of 0.63 between the three annotators, indicating good annotator agreement. \cref{fig:decision-tree} shows the decision tree used by annotators. \texttt{FP} is used to indicate a parsing failure, where the model failed to output a prediction in the expected format (ending in a quotation mark).

\subsection{Results for Other Models}\label{sec:other-models}
We repeat experiments for target language pairs \texttt{xx--en} for Apertus 8B \cite{DBLP:journals/corr/abs-2509-14233}, and OLMo-2 7B \cite{walsh2025} (henceforth simply Apertus and OLMo-2, respectively). Apertus is a recently released multilingual model, including 1,811 languages in its pretraining data. OLMo-2 was trained on Dolma, an English corpus \cite{soldaini-etal-2024-dolma}, hence, we do not list it in our discussion of available pretraining checkpoints in \cref{sec:related-work}. Despite this, OLMo-2 is known to have some multilingual capabilities (cf. \citealp{lee-etal-2025-controlling}), likely due to the ``accidental'' inclusion of multilingual data in its pretraining corpus \cite{blevins-zettlemoyer-2022-language}. For consistency with the main results, we patch from layer 10 for these models, even though they are deeper, i.e.\ have more layers. Since we patch all subsequent layers as well, patching earlier for the deeper models is a conservative choice; we see in preliminary experiments in \cref{sec:tech} that trends are fairly robust to layer choice.

\begin{figure*}
\centering
\includegraphics[width=\linewidth]{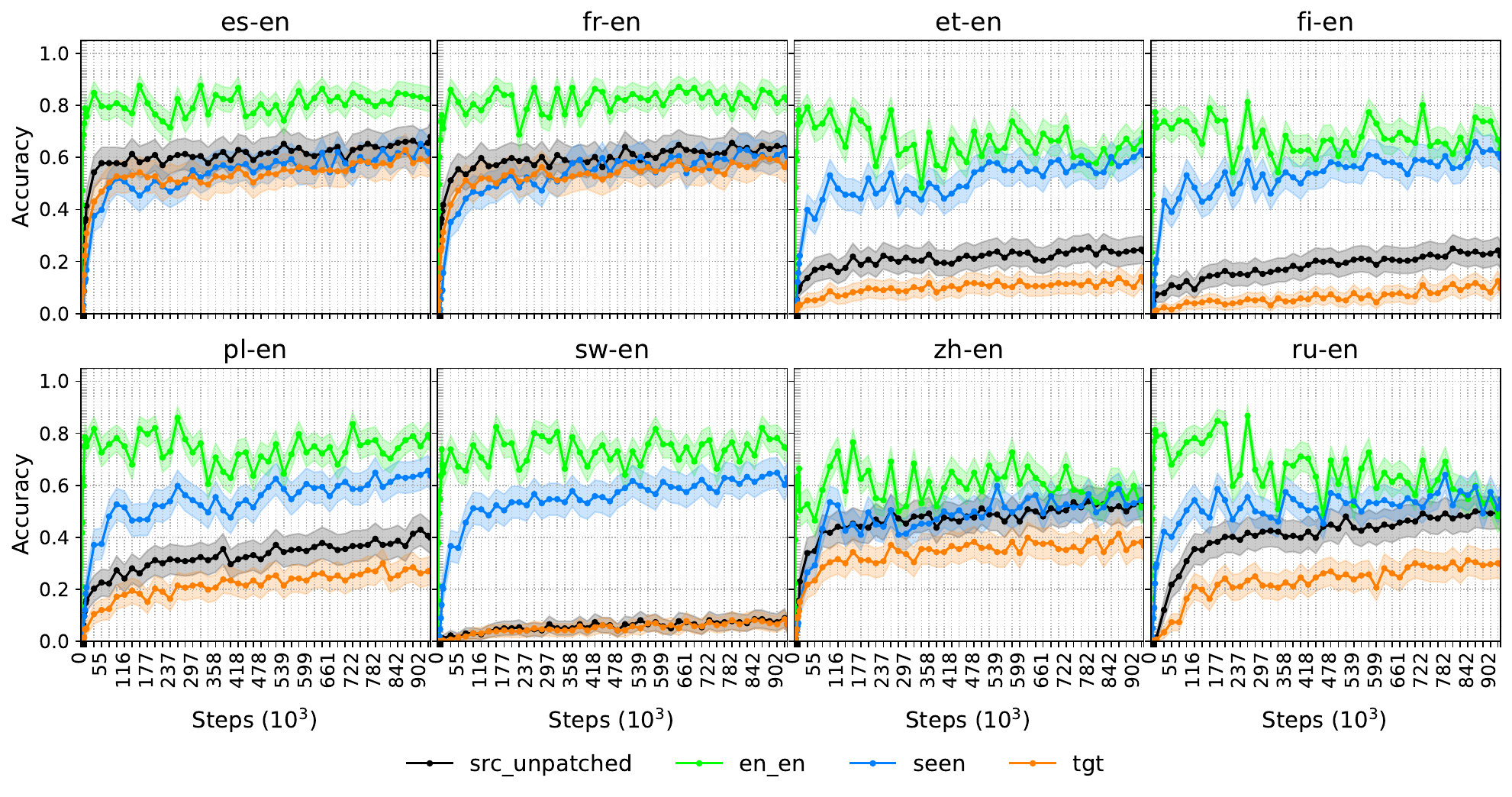}
\caption{Mean word-level translation accuracy over checkpoints of OLMo-2 7B under patching for target output language \texttt{en}. We show 95\% CI over 256 samples.}
\label{fig:olmo}
\end{figure*}

\begin{figure*}
\centering
\includegraphics[width=\linewidth]{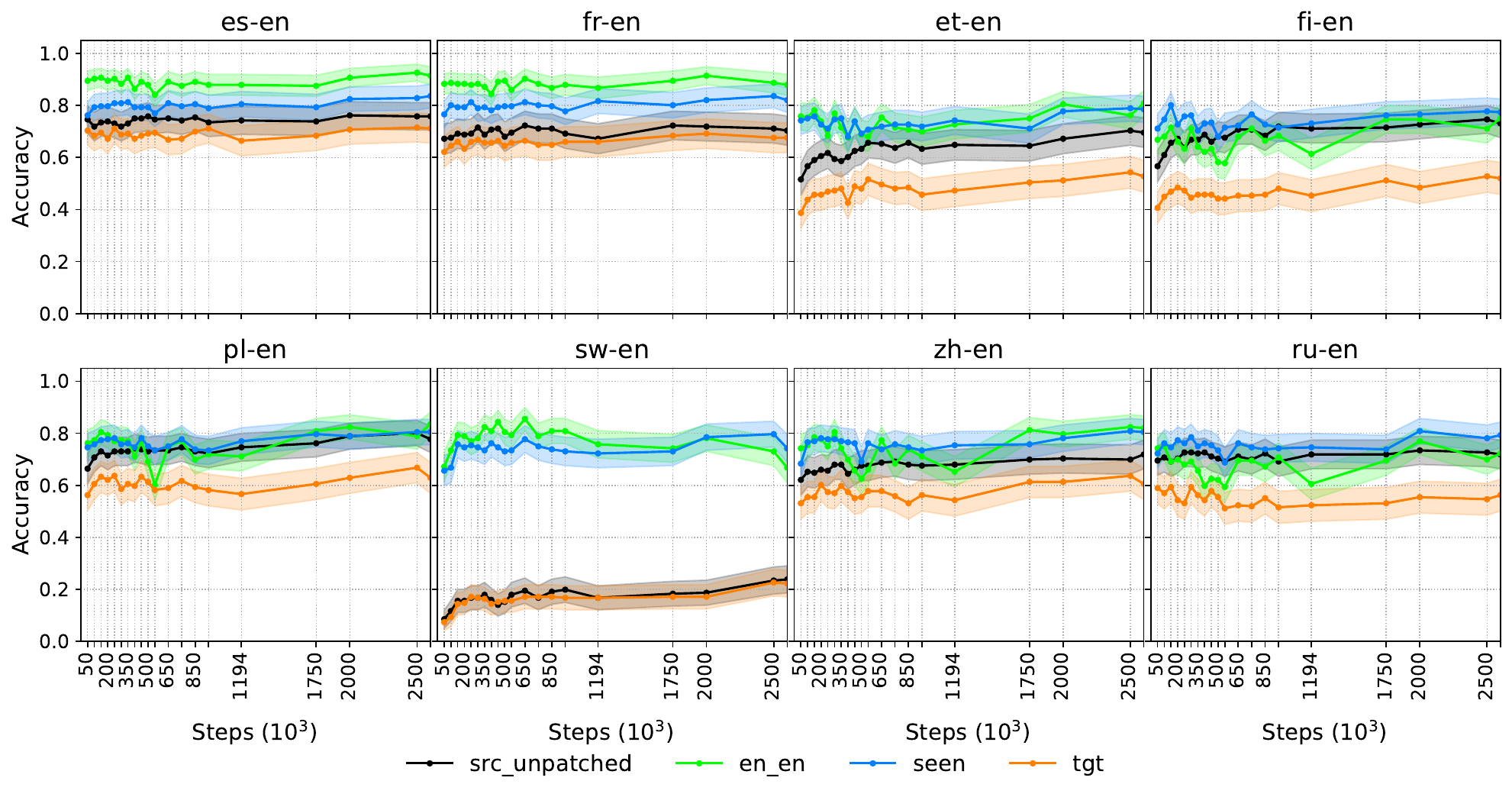}
\caption{Mean word-level translation accuracy over checkpoints of Apertus 8B under patching for target output language \texttt{en}. We show 95\% CI over 256 samples.}
\label{fig:apertus}
\end{figure*}

For both models, we see evidence of a shared concept space, as seen in the successful translation under the \texttt{seen} setting (see \cref{fig:olmo} for OLMo-2, and \cref{fig:apertus} for Apertus). This is more surprising for OLMo-2, which is trained on English data. However, we do see this English-centric training reflected in the \texttt{en\_en} performance, which is generally stronger than that under \texttt{seen} setting, especially in the beginning of training. This is in contrast to EuroLLM and Apertus, where the difference in \texttt{en\_en} and \texttt{seen} is much less pronounced.

Though we commend the rare release of multilingual pretraining checkpoints by the developers of Apertus, the first available checkpoint is already at 210B tokens consumed (compared to roughly 48B at the first checkpoint of EuroLLM, and 1B at the first checkpoint of OLMo-2). This low granularity of early checkpoints appears to mask the development of shared concept spaces \textemdash at the first checkpoint cross-lingual concept patching is successful, and does not improve much during training. However, the unpatched translation also does not improve much over checkpoints (with the exception of \texttt{en}, \texttt{fi}, \texttt{sw}, which are presumably lower-resourced), indicating that the model has already learned to translate our concepts by the first checkpoint. For OLMo-2, there is an increase in translation accuracy under \texttt{seen} over training checkpoints, suggesting that cross-lingual alignment increases over training. We hypothesize that this reflects the small amounts of multilingual data in OLMo-2's pretraining data, such that alignment may occur much later, as multilingual data is consumed by chance.

Overall, these results suggest that our main finding applies to other models: shared spaces develop early, in particular for models trained with multilingual data. Furthermore, multilingual pretraining data appears to drive development of this space, such that patching from \texttt{seen} is more successful for Apertus and EuroLLM than it is for OLMo-2, where the latter is trained on much fewer multilingual data.


\begin{figure*}
\centering
\includegraphics[width=\linewidth]{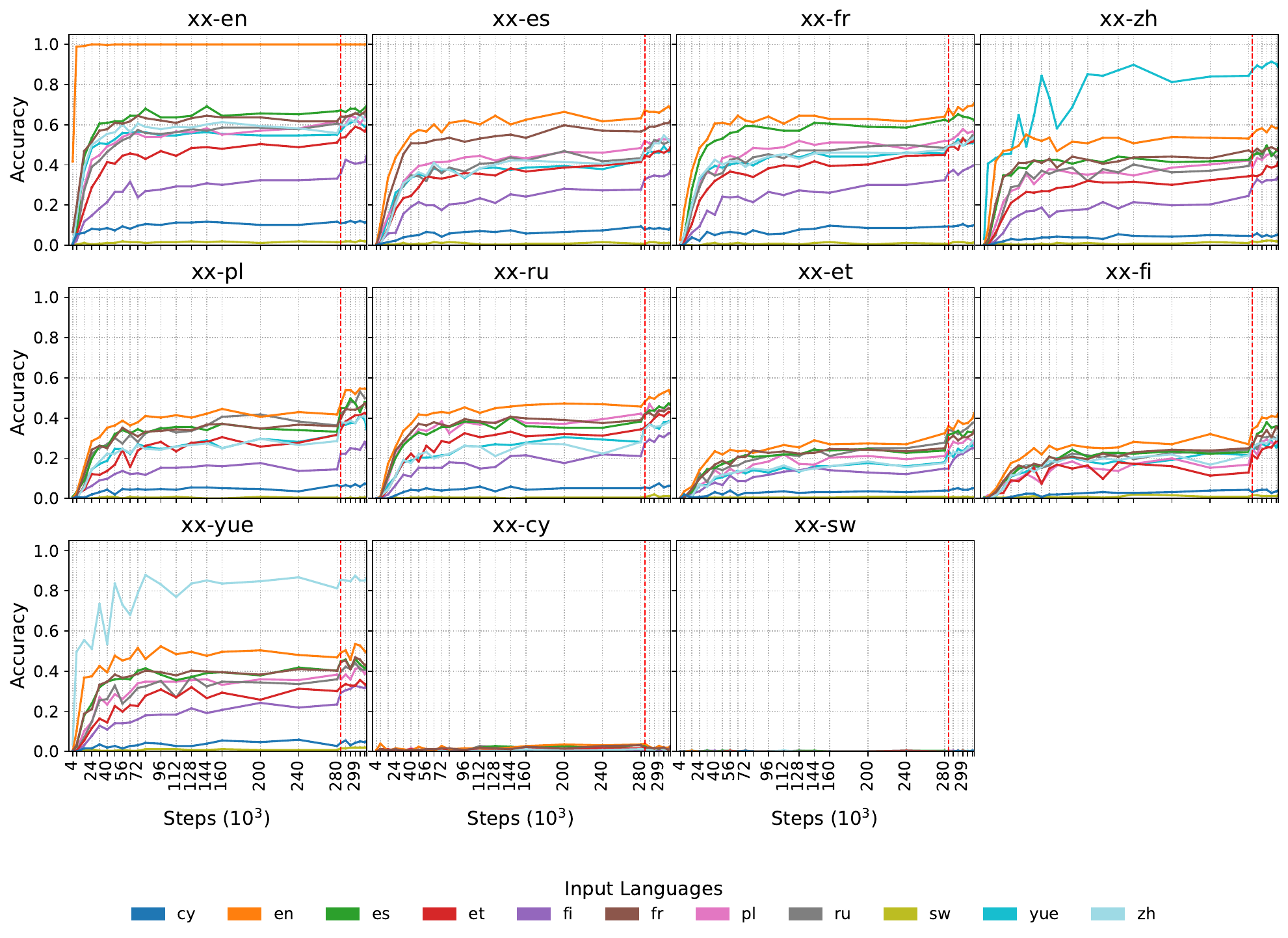}
\caption{Mean word-level translation accuracy over checkpoints for source prompts used for patching, grouped by output language for all language pairs. The red dotted line indicates the start of phase two of EuroLLM's training.}
\label{fig:full-translation}
\end{figure*}
\begin{figure*}
\centering
\includegraphics[width=\linewidth]{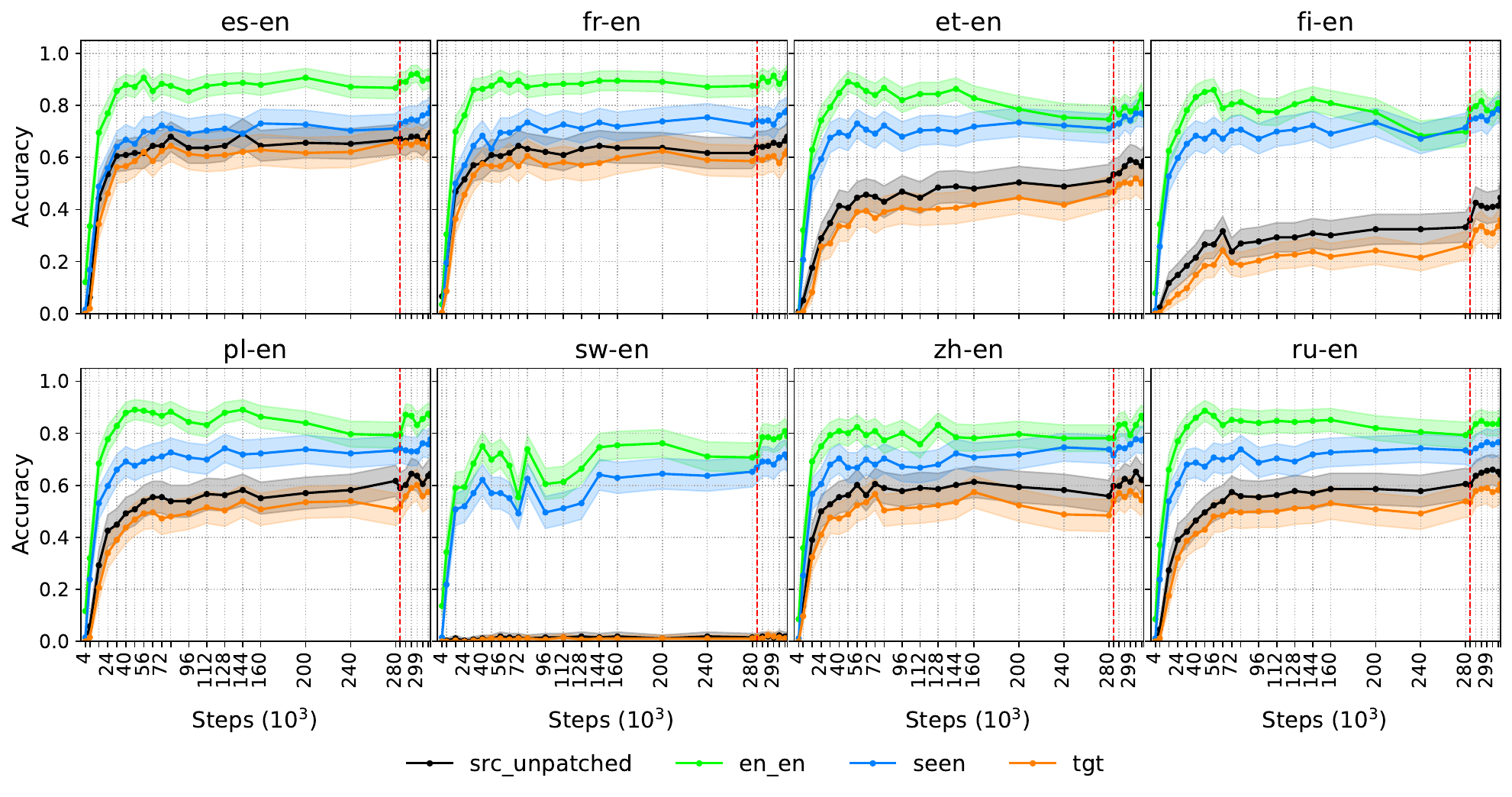}
\caption{Mean word-level translation accuracy over checkpoints under patching for target output language \texttt{en}. The red dotted line indicates the start of phase two of EuroLLM's training. We show 95\% CI over 256 samples.}
\label{fig:obj-patch-full-en}
\end{figure*}
\begin{figure*}
\centering
\includegraphics[width=\linewidth]{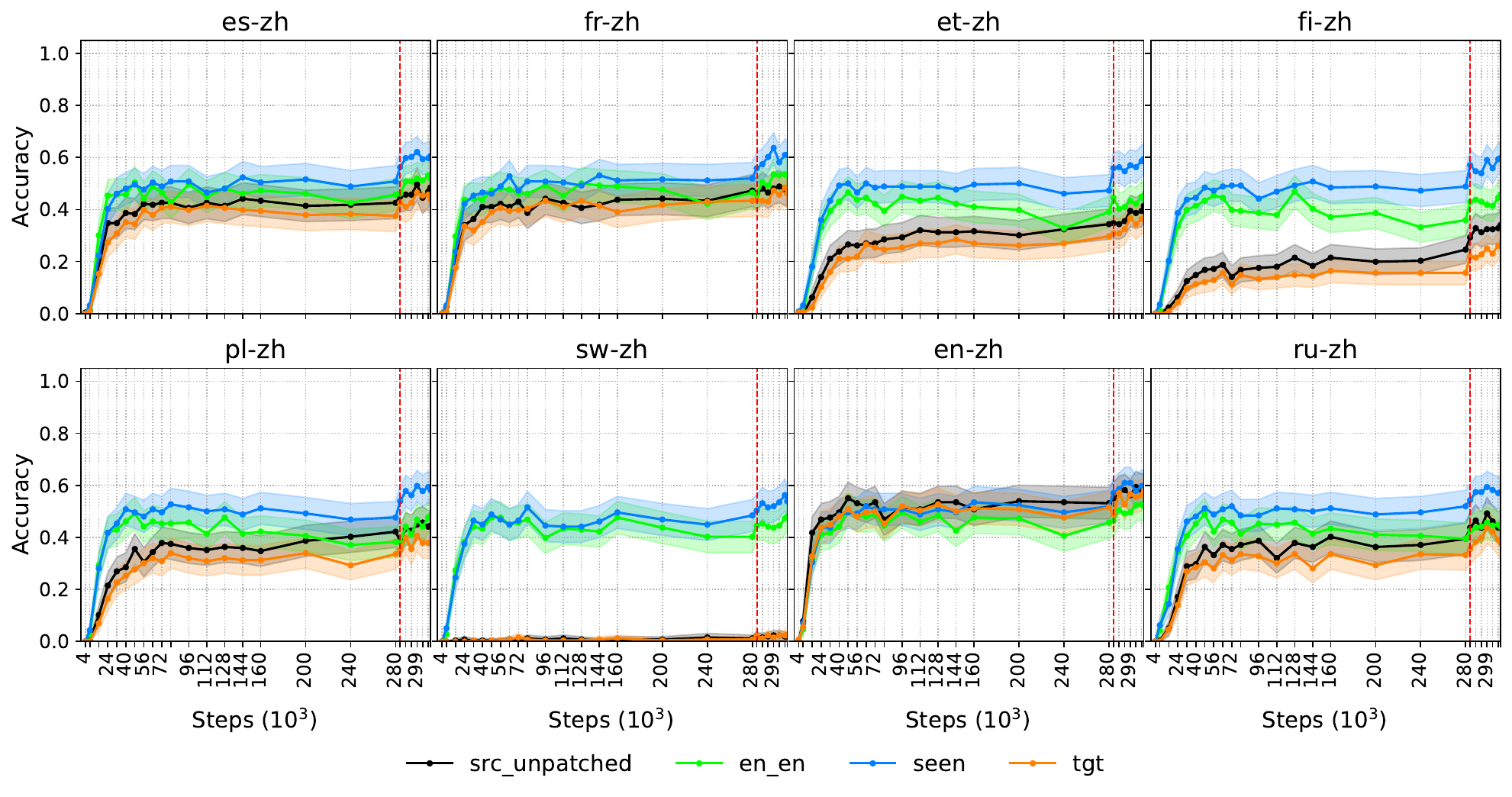}
\caption{Mean word-level translation accuracy over checkpoints under patching for target output language \texttt{zh}. The red dotted line indicates the start of phase two of EuroLLM's training. We show 95\% CI over 256 samples.}
\label{fig:obj-patch-full-zh}
\end{figure*}
\begin{figure*}
\centering
\includegraphics[width=\linewidth]{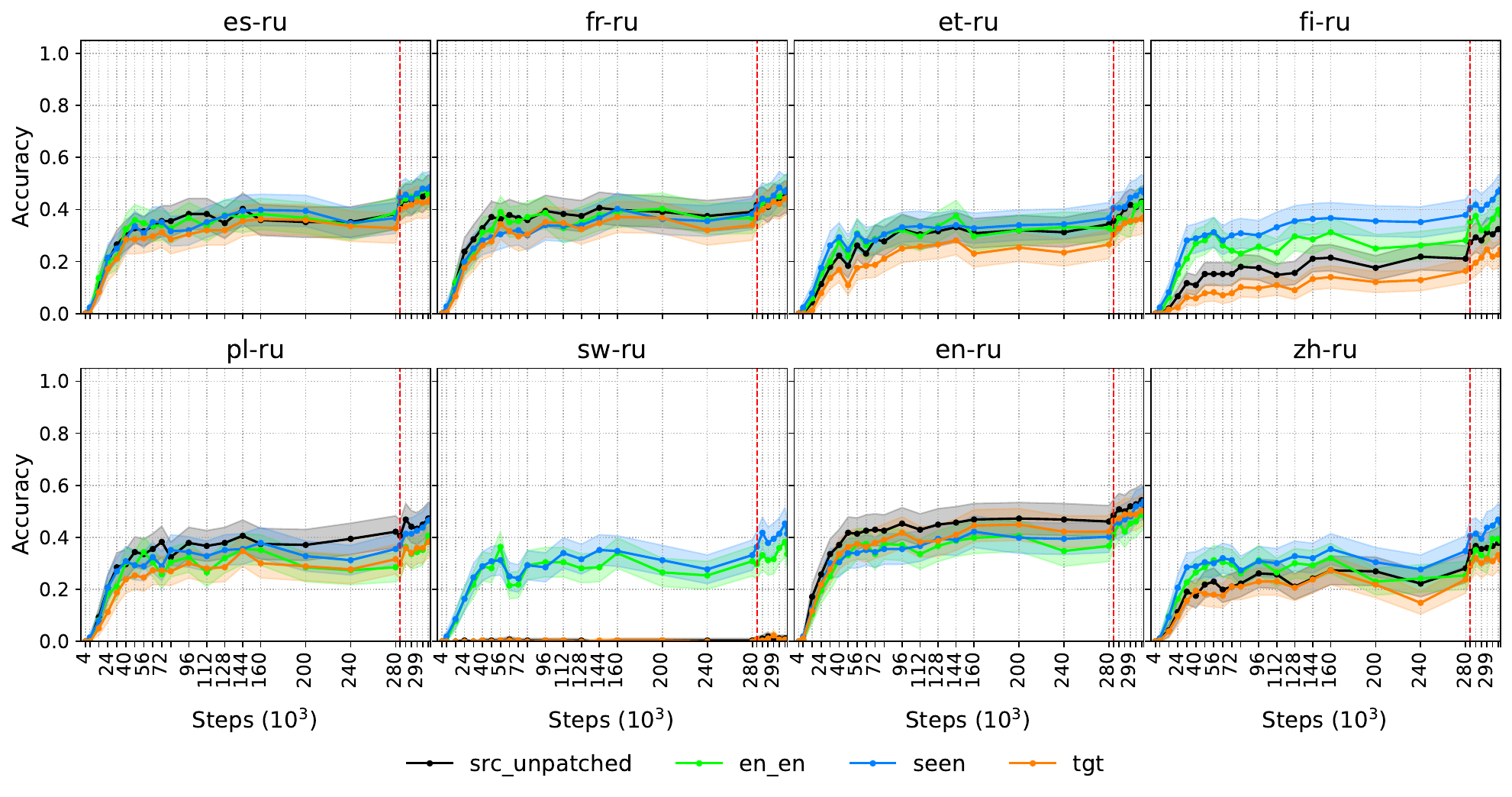}
\caption{Mean word-level translation accuracy over checkpoints under patching for target output language \texttt{ru}. The red dotted line indicates the start of phase two of EuroLLM's training. We show 95\% CI over 256 samples.}
\label{fig:obj-patch-full-ru}
\end{figure*}

\begin{figure*}[t]
\centering
\includegraphics[width=\linewidth]{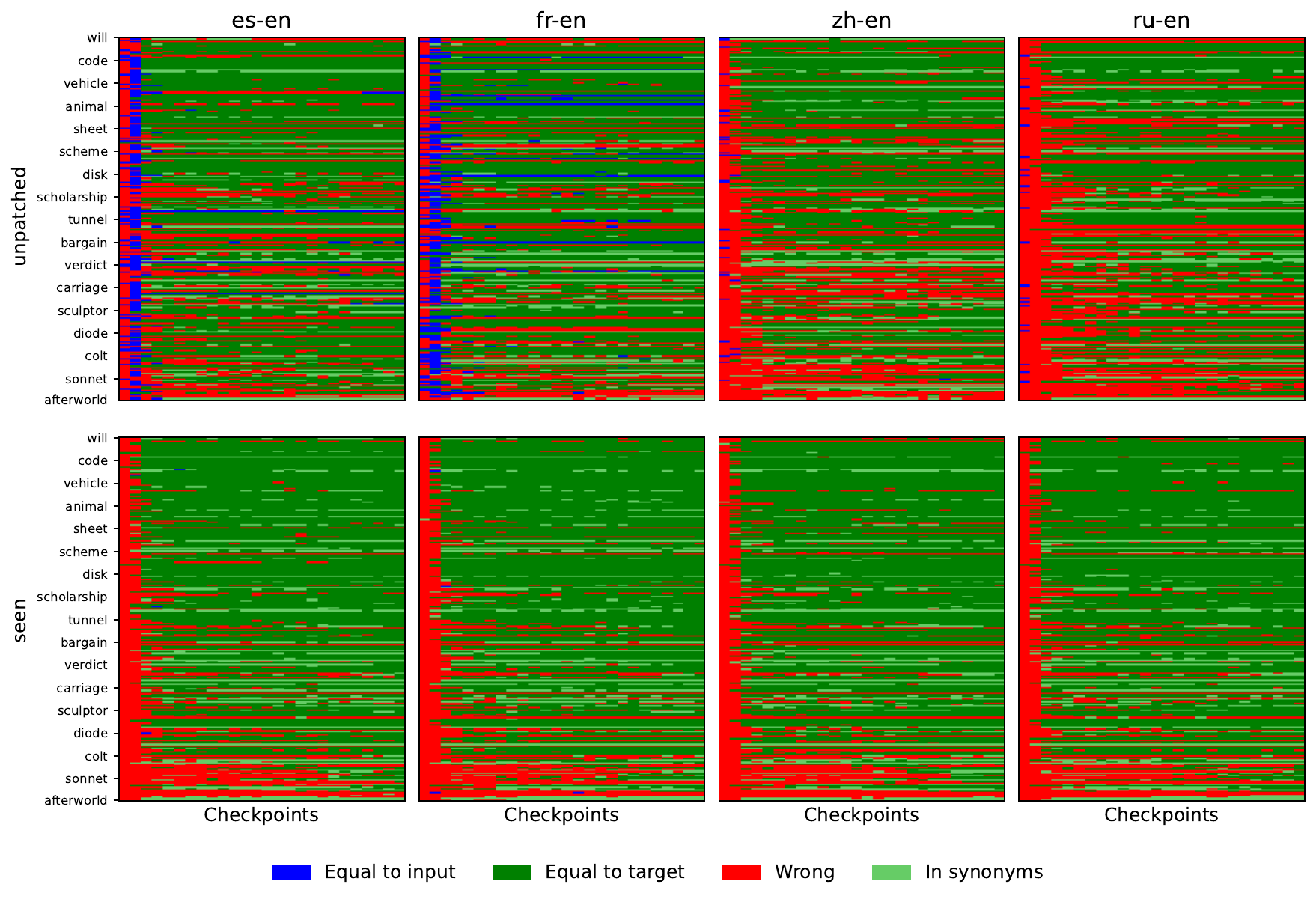}
\caption{Translation categorical maps under the unpatched and \texttt{seen} setting for \texttt{es--en}, \texttt{fr--en}, \texttt{zh--en}, \texttt{ru--en}  sorted by target word frequency. Each row represents a target concept, with a selection of concepts shown as ticks. Each column represents one checkpoint.}
\label{fig:seen-unpatched-freq}
\end{figure*}

\begin{figure*}
\centering
\includegraphics[width=\linewidth]{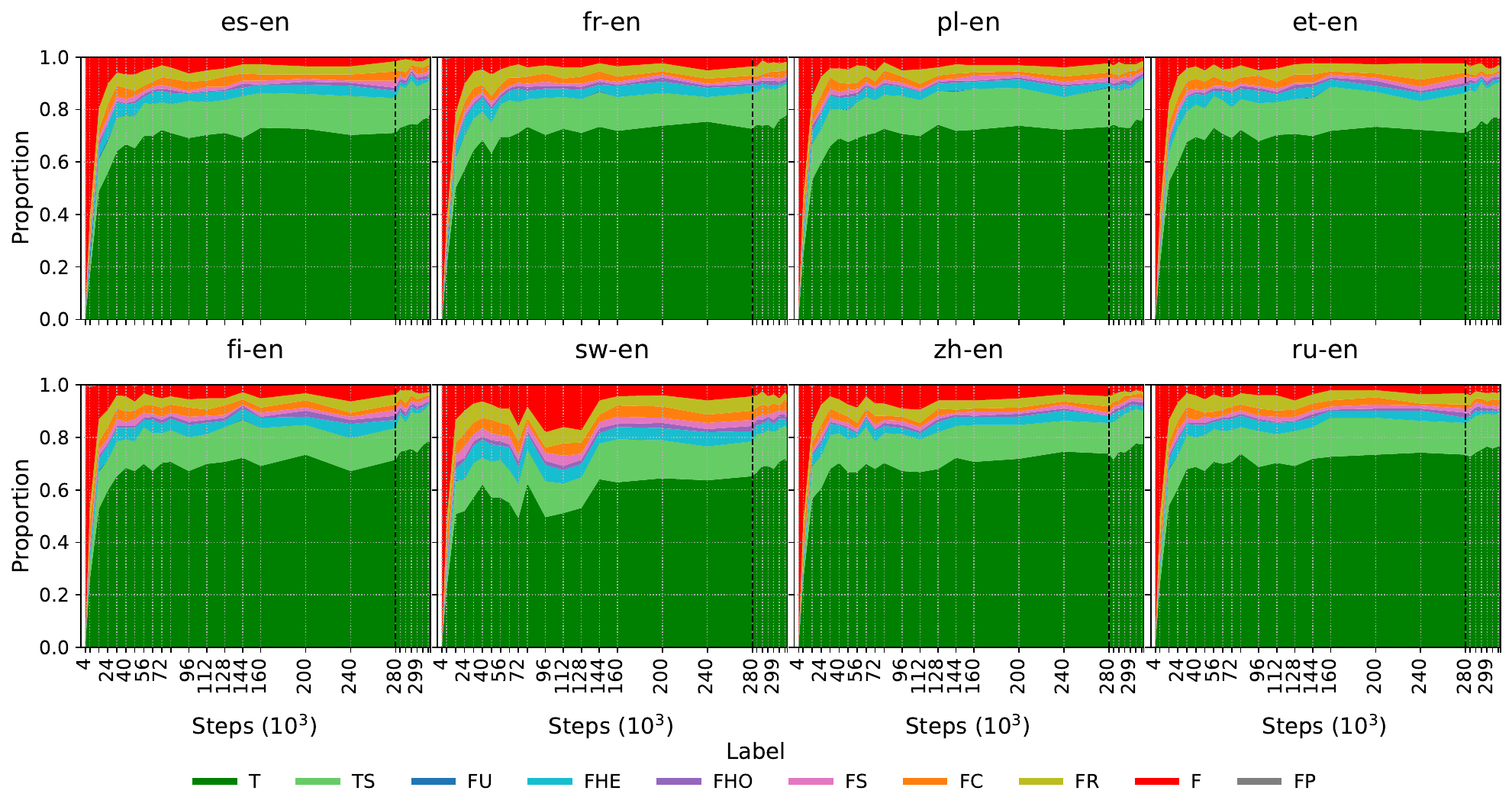}
\caption{Area grid of label distribution for outputs under the \texttt{seen} patching setting for target output language \texttt{en}. The black dotted line indicates the start of phase two of EuroLLM's training.}
\label{fig:obj-patch-area-full-en}
\end{figure*}
\begin{figure*}
\centering
\includegraphics[width=\linewidth]{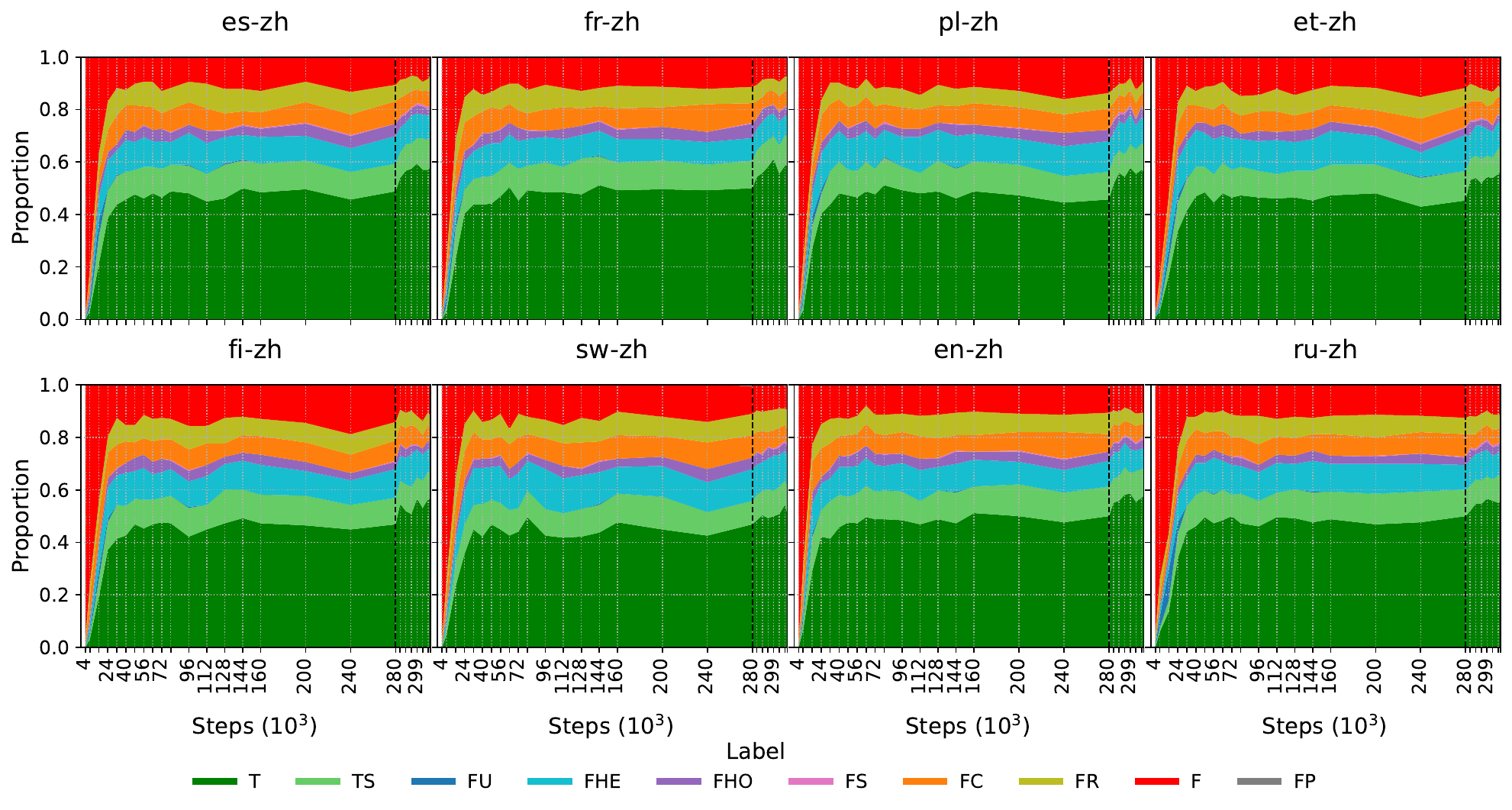}
\caption{Area grid of label distribution for outputs under the \texttt{seen} patching setting for target output language \texttt{zh}. The black dotted line indicates the start of phase two of EuroLLM's training.}
\label{fig:obj-patch-area-full-zh}
\end{figure*}
\begin{figure*}
\centering
\includegraphics[width=\linewidth]{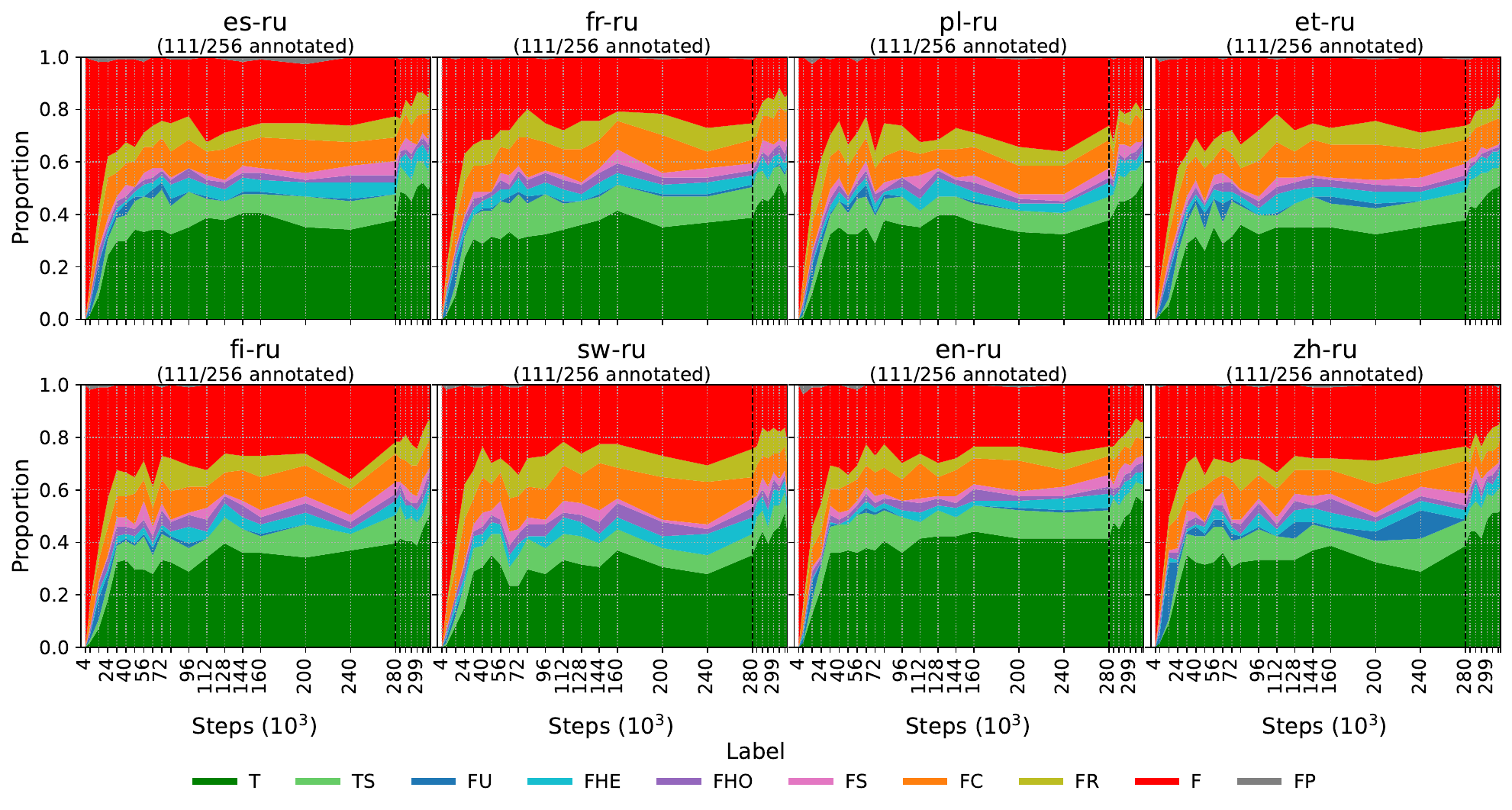}
\caption{Area grid of label distribution for outputs under the \texttt{seen} patching setting for target output language \texttt{ru}. We show only the subset of outputs that were labeled (corresponding to 111/256 concepts). The black dotted line indicates the start of phase two of EuroLLM's training.}
\label{fig:obj-patch-area-full-ru}
\end{figure*}
\end{document}